\documentclass[sigconf, 9pt]{acmart}

\renewcommand\footnotetextcopyrightpermission[1]{} 
\usepackage{booktabs} 
\usepackage[english]{babel}
\usepackage{blindtext}
\usepackage{graphicx}
\usepackage{amsmath}

\usepackage{paralist}
\usepackage{subfigure}
\usepackage{diagbox}
\usepackage{caption}
\usepackage{parskip}
\usepackage{xcolor}
\usepackage[ruled,vlined]{algorithm2e}

\begin{document}

\title{\name: Opportunistic Enhancement for Edge Video Analytics}


\author{Yan Lu$^*$, Shiqi Jiang$^\dagger$, Ting Cao$^\dagger$, Yuanchao Shu$^\dagger$}
\affiliation{$^*$New York University, $^\dagger$Microsoft}
\email{jasonengineer@hotmail.com, {shijiang, ting.cao, yuanchao.shu}@microsoft.com}

\begin{abstract}
Edge computing is being widely used for video analytics. To alleviate the inherent tension between accuracy and cost, various video analytics pipelines have been proposed to optimize the usage of GPU on edge nodes. Nonetheless, we find that GPU compute resources provisioned for edge nodes are commonly under-utilized due to video content variations, subsampling and filtering at different places of a pipeline. As opposed to model and pipeline optimization, in this work, we study the problem of opportunistic data enhancement using the non-deterministic and fragmented idle GPU resources. In specific, we propose a task-specific discrimination and enhancement module and a model-aware adversarial training mechanism, providing a way to identify and transform low-quality images that are specific to a video pipeline in an accurate and efficient manner. A multi-exit model structure and a resource-aware scheduler is further developed to make online enhancement decisions and fine-grained inference execution under latency and GPU resource constraints. Experiments across multiple video analytics pipelines and datasets reveal that by judiciously allocating a small amount of idle resources on frames that tend to yield greater marginal benefits from enhancement, our system boosts DNN object detection accuracy by $7.3-11.3\%$ without incurring any latency costs. 
\end{abstract}

%
%
\newcommand{\para}[1]{\medskip\noindent\textbf{#1}}

\newcommand{\name}[0]{{\sc Turbo}\xspace}

\newcommand{\ie}{{\it i.e.,}\xspace}
\newcommand{\eg}{{\it e.g.,}\xspace}
\newcommand{\ys}[1]{\colorbox{red}{yuanchao:} \textcolor{cyan}{#1}}
\newcommand{\ysedit}[1]{{\leavevmode\color{cyan}{#1}}}
\newcommand{\yan}[1]{\colorbox{olive}{yan:} \textcolor{olive}{#1}}
\newcommand{\sq}[1]{\colorbox{magenta}{shiqi:} \textcolor{orange}{#1}}
\newcommand{\sqedit}[1]{{\leavevmode\color{magenta}{#1}}}
\newcommand{\todo}[1]{\colorbox{blue}{todo:} \textcolor{blue}{#1}}
\newcommand{\ting}[1]{\colorbox{brown}{ting:} \textcolor{brown}{#1}}
\maketitle

\section{Introduction}
\label{sec:intro}


Video analytics has drawn a significant attention over the past couple years due to the growing presence of cameras and rapid developments on artificial intelligence. In order to preserve privacy and lower the total cost of ownership for video analytics, edge compute devices are predominantly used at customer's premises for video ingestion and processing~\cite{wu2021pecam, ran2018deepdecision, yi2017lavea, Reducto, eaar2019, vigil2015, glimpse2015}. 

Edge devices are known to be resource-constrained. Over the years, a considerable amount of literature has been published on the design and implementation of efficient edge video analytics. Examples include but are not limited to cascaded and adaptive analytics pipeline~\cite{enomoto2021cascadeInference,jiang2018chameleon,ga19mobisys,rocketblog,distream2020sensys}, multi-capacity neural networks~\cite{fang2021flexdnn,han2021dynamic,fangzeng2018nestdnn}, memory-efficient deep neural network (DNN) inference~\cite{Padmanabhan21hotedgevideo}, low-cost analytics across cameras~\cite{jain20sec} and hierarchical clusters~\cite{hung2018videoedge}.
In contrast to the plethora of research on the optimizations of video analytics pipelines (VAPs) and DNNs, in this paper, we seek to answer the question that \emph{given an optimized VAP}, how idle compute resources on edge, if present, can be harnessed to further improve the overall analytics accuracy.

The rationale behind this question is two-fold. First, by studying canonical VAPs on real-world datasets, we noticed that there is a decent amount of idle GPU compute resources on the edge due to video content changes and widely-used subsampling techniques~\cite{jiang2018chameleon,hung2018videoedge,ga19mobisys,Reducto,rocketblog}. For instance, in a cascaded analytics pipeline, heavyweight DNN is called upon only when a lightweight CPU-based background subtraction module detects motion in certain areas, resulting in less but video-dependent and fluctuating GPU usage. Similarly, a vehicle counting and recognition pipeline could generate much less DNN inference requests at times of low traffic volume. The same observation holds true to a wide range of VAPs given that edge machine is commonly used to process multiple camera streams and tends to be provisioned for scenarios of the worst-case workload. Second, we found that despite of decent overall accuracy provided by VAPs on target video inputs, there always exists a small portion of frames where VAPs perform poorly. This can be due to many reasons, including the low quality of the image (\eg occlusion, blur, low lighting), and the lack of representative training data for the DNN. Regardless of the cause, analytics accuracy could be largely improved from effective enhancements on such hard samples.

To reap the benefits of idle GPU resources and further improve the performance of an existing VAP, we introduce \name, an opportunistic enhancement framework which \emph{selectively} enhances incoming frames based on GPU resource availability and characteristics of the DNN model used in a VAP. 
Design of \name, however, faces three challenges. First, it is non-trivial to reliably and efficiently identify frames that tend to yield inferior performance on downstream DNNs. The reason is simply because inference performance depends both on frame contents and on the DNN used in a VAP. For example, an object that is easily recognizable by a DNN detector could become ambiguous in a couple frames when lighting condition changes. Likewise, the definition of \emph{hard} might vary significantly between a YOLOv3~\cite{yolov3} model pre-trained on COCO dataset~\cite{coco2014} and a Faster-RCNN~\cite{fasterrcnn2015} model pre-trained on a private dataset. Second, it is technically challenging to improve the performance of an existing DNN on hard samples without sacrificing its accuracy on relatively easy ones. End-to-end model optimizations (\eg retraining or fine-tuning the entire model for hard samples) could lead to overfitting or bias, and is also prohibitively expensive in terms of both compute and annotation cost. At times model adaptation and retraining could even become infeasible when proprietary DNNs and techniques (\eg third-party software, specialized accelerators) are used.
Third, idle GPU resources from running VAPs are \emph{non-deterministic} and \emph{fragmented}. Hence, enhancement at runtime requires the awareness of resource availability as well as an elastic and fine-grained execution mechanism. 

\begin{figure*}[!t]
    \centering
    \includegraphics[width=0.95\linewidth]{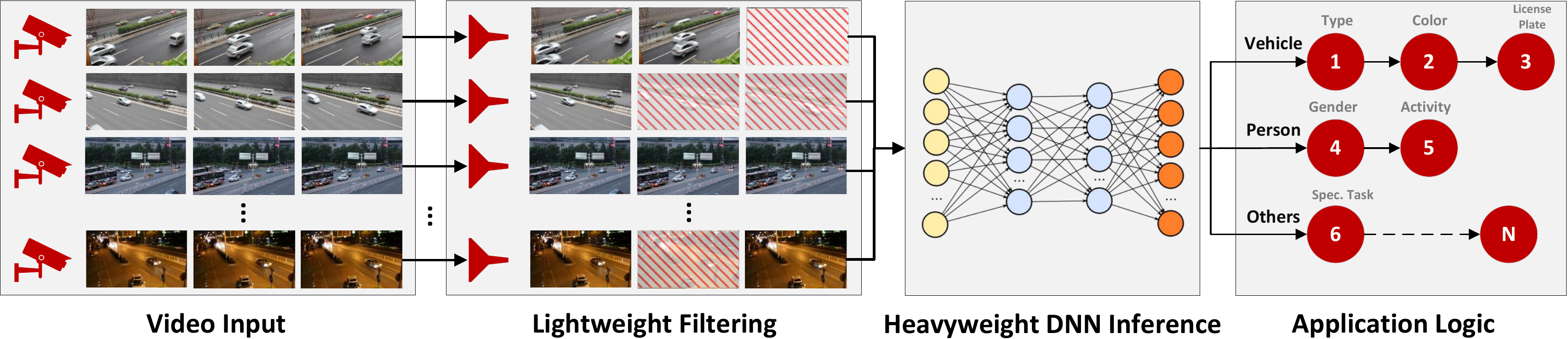}
    \caption{Illustration of a sample cascaded edge video analytics pipeline. In this paper, we use the object detection as an example of the heavyweight DNN tasks.}
    \label{fig:illustration_vap}
\end{figure*}

In \name, we tackle these challenges by making the following three contributions.

\begin{itemize}

    \item We propose a task-specific discrimination and enhancement module based on generative adversarial networks (GAN). The module is trained by a novel model-aware adversarial training mechanism, which as a result, provides a \emph{discriminator} that effectively identifies hard samples for a particular DNN, and a \emph{generator} that makes image inputs more amenable to the downstream DNN in an efficient manner. 
    
    
    \item We devise an enhancement execution module, achieved by an elastic structure design of the GAN model and a resource-aware scheduler, to best utilize the fragmented GPU compute resources. Specifically, the module maximizes the overall analytics accuracy by running a pre-trained multi-exit GAN model at different enhancement levels on selected frames under given latency and resource constraints. 
    
    

    \item We fully implement our solution and evaluate it on two large-scale real-world video datesets. Results from three video analytic pipelines show that without incurring any latency cost, average analytic accuracy improves by $9.0\%$, $11.3\%$, and $7.2\%$ for three different detection models from idle resource harvesting.
    
\end{itemize}

In what follows, we use the object detection, a pivotal component in various video analytics systems, as a canonical application to motivate and describe the design of \name. \name can be easily extended to other kinds of heavyweight video DNN workloads as we only rely on DNN output and do not make any assumption on the inner workings of the model. 
%

\section{Motivation and Background}
\label{sec:motivation}
In this section, we present the opportunities and challenges in the opportunistic enhancement for edge video analytics.

\subsection{Edge Video Analytics Pipelines}
\label{subsec:edge_vap}
\begin{figure}[!t]
    \subfigure[Vigil]
    {
        \includegraphics[width=0.475\linewidth]{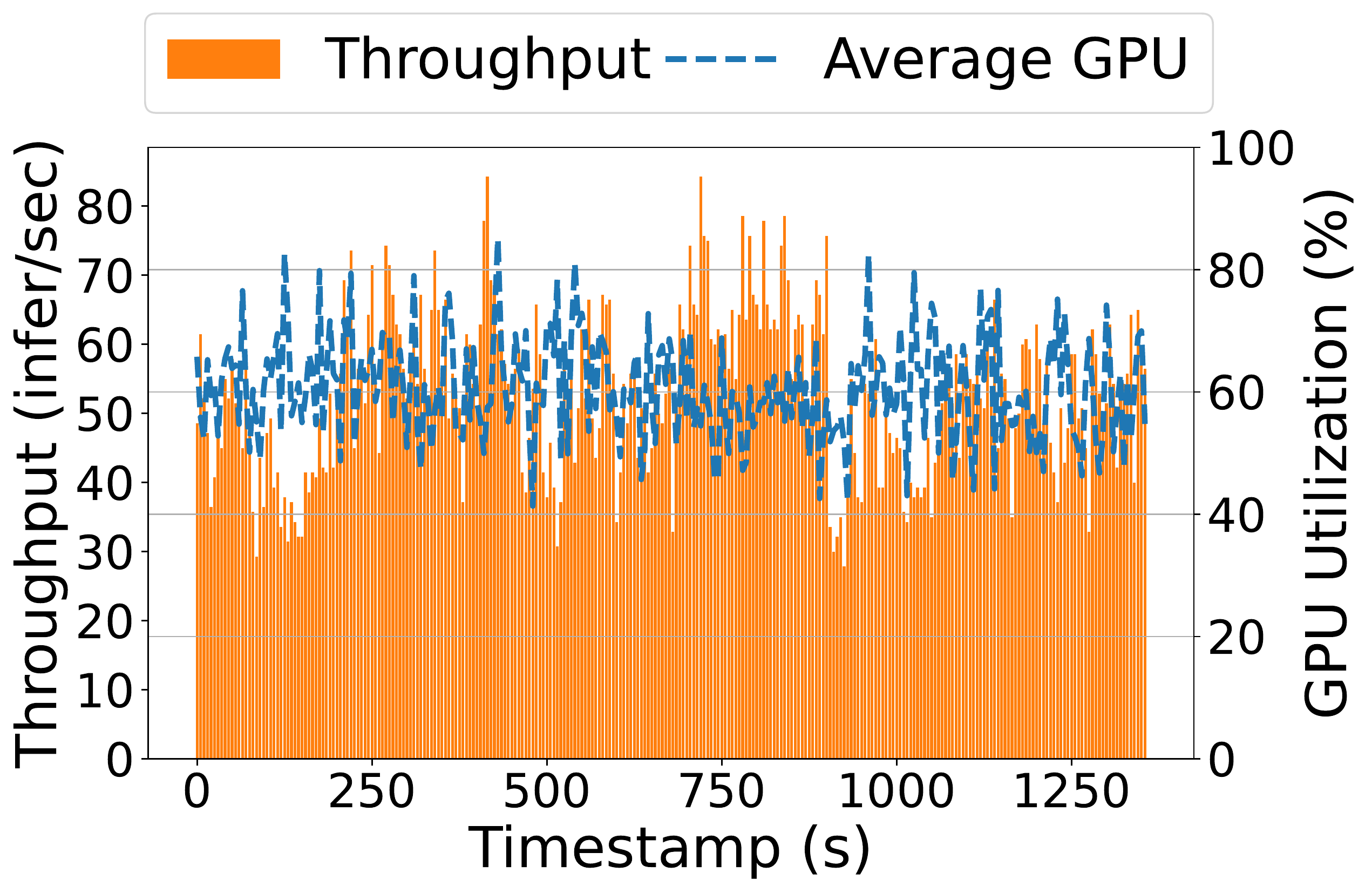}
        \label{subfig:dynamic_vigil}
    }
    \hfill
    \subfigure[Glimpse]
    {
        \includegraphics[width=0.475\linewidth]{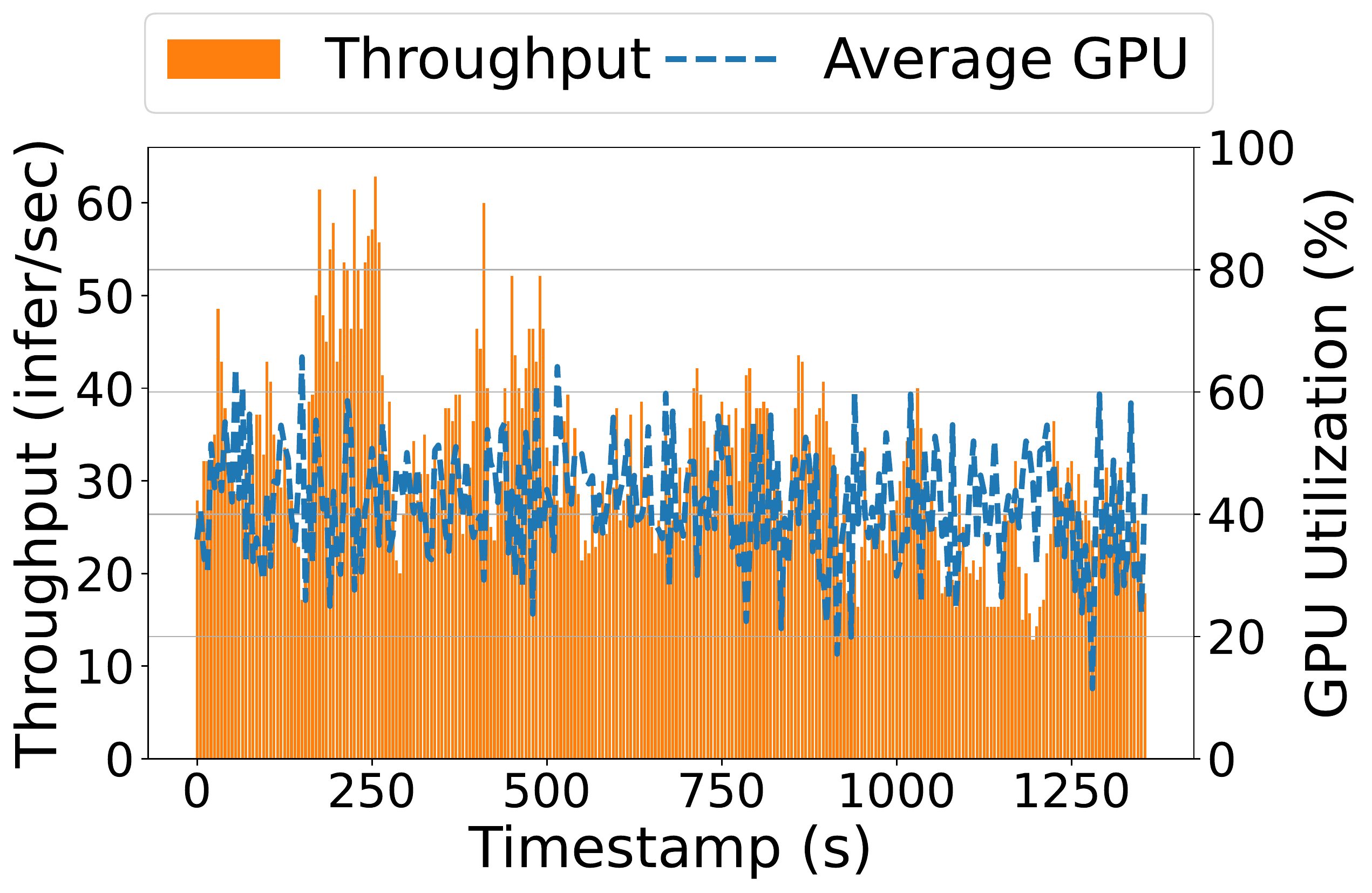}
        \label{subfig:dynamic_glimpse}
    }
    \caption{Dynamic video analytics workloads on an edge device shared by multiple streams.}
    \label{fig:dynamic_workload}
\end{figure}

Edge devices are being used increasingly for video analytics. Given the nature of limited compute and network resources, video analytics task typically uses a cascaded pipeline which consists of a series of modules on the decoded frames of the video stream. Fig.~\ref{fig:illustration_vap} demonstrates a VAP, where multiple cameras are connected to an edge node, and on it downstream modules like DNN-based object detection are performed. 
Before reaching heavyweight DNNs, video frames are typically processed using techniques like temporal pruning (\eg sub-sampling based on pixel differences between frames), spatial pruning (\eg region cropping and background extracting), and model pruning (\eg model specialization and cascading)~\cite{measurement2021xiao}. Such modules result in more efficient but dynamic GPU usage which are content-dependent. 

To examine the performance of edge VAPs and their corresponding resource utilization in real deployments, we conduct a measurement study with two canonical pipelines, Glimpse~\cite{glimpse2015} and Vigil~\cite{vigil2015}. Glimpse uses temporal pruning and sends frames to the downstream object detection model only when movements are detected between two frames. Vigil, on the other hand, adopts model pruning and sends out only images that contain objects detected by a cheap local model. We execute these two pipelines on UA-DTRAC~\cite{CVIU_UA-DETRAC}, a traffic video dataset with rich annotations. In all experiments, we use EfficientDet-D0~\cite{efficientdet2020} as the object detection model, and process 4 video streams simultaneously 
on an Azure Stack Edge Pro~\cite{azure_stack_edge} equipped with a NVIDIA Tesla T4 GPU~\cite{nvidia_t4}. 
We use the Streaming Multiprocessor (SM) Activity reported by NVIDIA DCGM~\cite{dcgm2021} to characterize the GPU utilization. SM activity is defined as the fraction of time at least one warp was active on a SM, averaged over all SMs. It is a finer-grained metric than \emph{nvidia-smi}'s GPU utilization number, which is the ratio of time the graphics engine is active.


\begin{figure}[!t]
        \centering
        \includegraphics[width=0.98\linewidth]{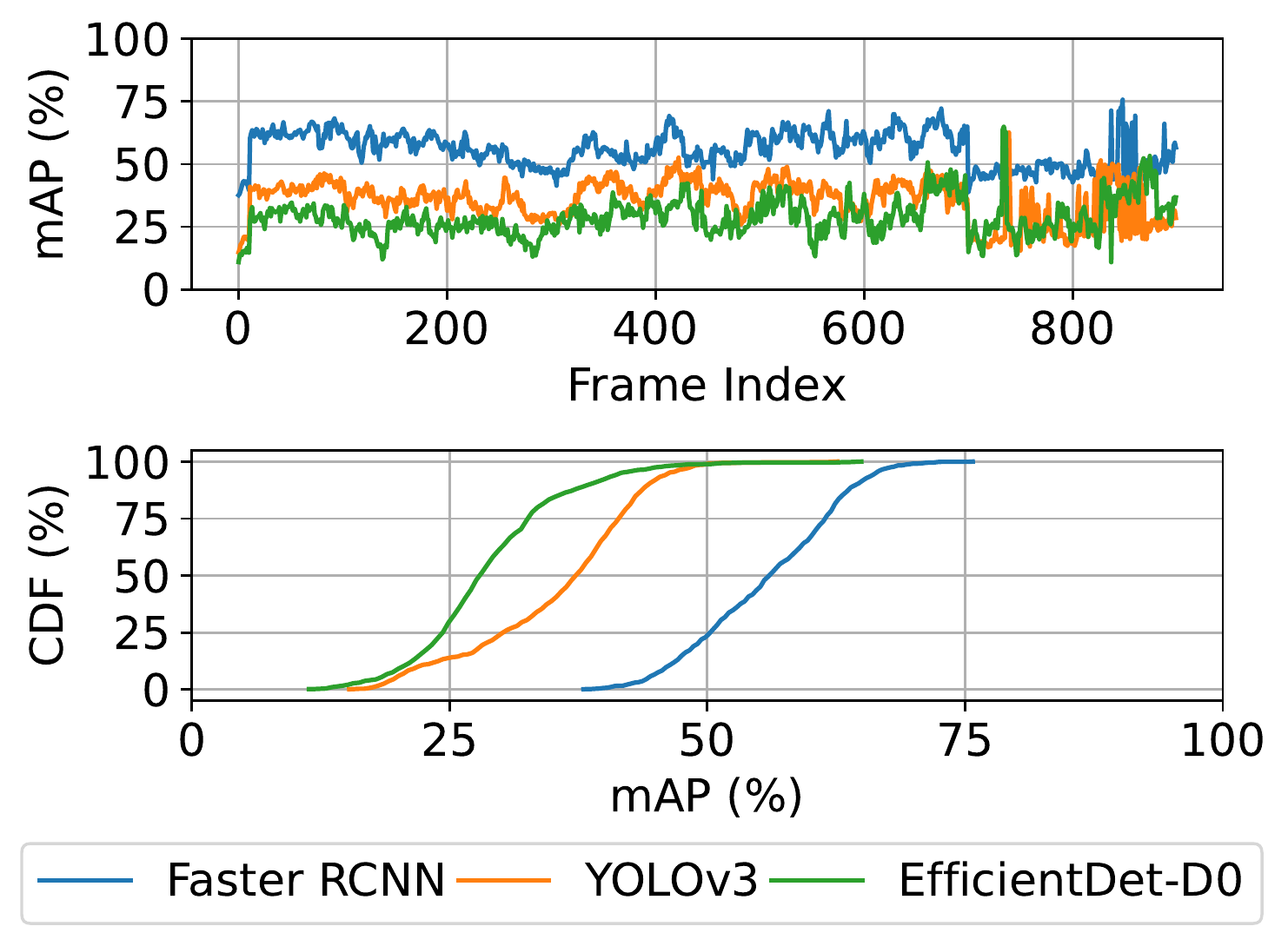}
        \caption{mAP of each individual frame from the selected video trace of UA-DTRAC using different models.}
        \label{fig:short_time_performance_jitters}
\end{figure}

Due to pruning, only a portion of frames are eventually sent to the GPU for processing. Fig.~\ref{fig:dynamic_workload} illustrates the actual workloads and the corresponding GPU utilization of a selected VAP. Overall, we observe that GPU throughput varies greatly over time, from $11$ infer/sec to $78$ infer/sec. In particular, for Glimpse, throughput goes beyond $50$ infer/sec for only $7.2\%$ of the time. Similarly, throughput of Vigil stays below $45$ infer/sec for $19.0\%$ of the time. 

Not surprisingly, the GPU usage also fluctuates due to workload variations. Specifically, there appears more than $43\%$ and $60\%$ idle resources on average for Vigil and Glimpse, respectively. Furthermore, the appearances of idle resources are non-deterministic and fragmented since they are highly related to video contents. It is a fleeting opportunity to harvest the idle resource and in turn improve the analytics accuracy. 


We also study the performance of different object detection models, including EfficientDet~\cite{efficientdet2020}, Faster-RCNN~\cite{fasterrcnn2015} and YOLOv3~\cite{yolov3}, on each individual frame of a selected trace from UA-DTRAC. From Fig.~\ref{fig:short_time_performance_jitters}, we notice that the mean averaged precision (mAP) varies dramatically over time. For Faster RCNN, while more than half of the frames (425) yield accuracy higher than $55.0\%$, due to low mAP scores on a small set of hard frames, the averaged mAP across all frames is only $52.7\%$. In fact, the averaged mAP of the bottom $5\%$ of the frames (45) is as low as $37.1\%$. Similar results are observed on both beefier and wimpier models. For example, the average mAP of YOLOv3 can be increased by $10.6\%$ if the mAP of the bottom $5\%$ frames is improved by $15.5\%$. 

\subsection{Challenges of Opportunistic Enhancement}




To improve the performance of a VAP on hard samples, one might employ a DNN model optimization approach by collecting all hard samples, annotating and using them to retrain or fine-tune the model. The method, however, falls short for two reasons. First, data collection and annotation process could be computationally expensive and the fragmented idle GPU resources makes the DNN model training challenging. Second, VAPs in real deployment might consist of black-box DNN models pre-trained on proprietary datasets. Without the details (\eg DNN architecture and weights), one can hardly update the model. In fact, even for a model well pre-trained and fine-tuned, there still exist hard samples and under-utilized compute resources given the nature of a pipeline and content variations. 

As such, we shift our focus to opportunistically enhancing input images of a video analytics system. Image enhancement has been extensively studied in both computer vision and systems communities~\cite{liang2021swinir,qin2020ffa,Zamir2021MPRNet,liang2021swinir,jiang2021enlightengan,Zero-DCE2020, EagleEye, Elf, MobiSR}. 
Enhancement methods, such as super resolution, deblurring and dehazing, look for ways to restore corrupted details in raw captured images. Intuitively, one can apply off-the-shelf image enhancements to improve the quality of frames for edge video analytics. 

\begin{figure}[t]
    \centering
    \includegraphics[width=0.95\linewidth]{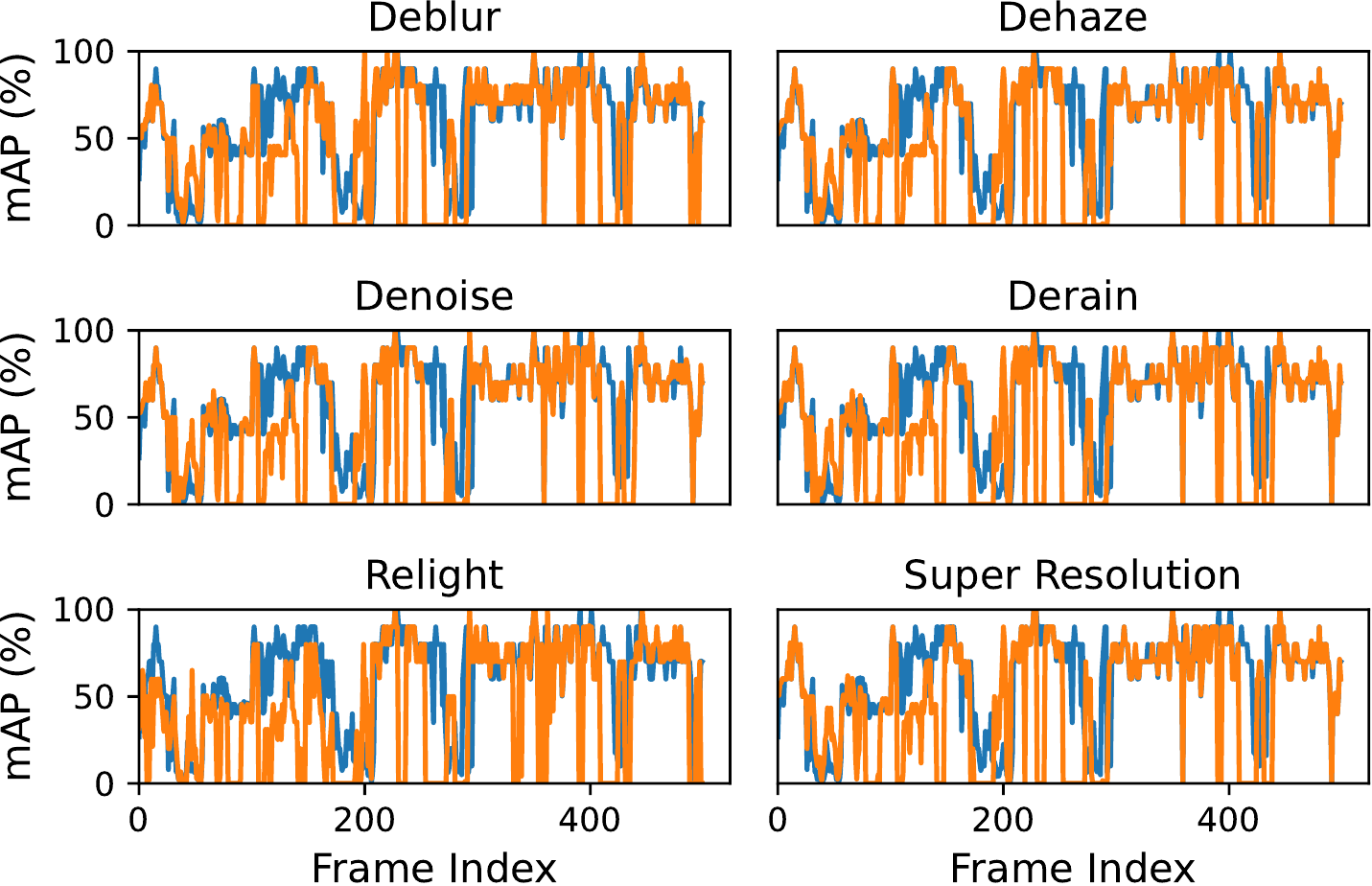}
    \caption{mAP on every raw \textit{(blue)} and enhanced \textit{(orange)} image of a selected video trace. }
    \label{fig:accuracy_per_image_with_ie}
\end{figure}

To this end, we select six state-of-the-art image enhancement methods, namely super resolution~\cite{liang2021swinir}, dehaze~\cite{qin2020ffa}, deblur~\cite{Zamir2021MPRNet}, denoise~\cite{liang2021swinir}, relight~\cite{jiang2021enlightengan, Zero-DCE2020} and derain~\cite{Zamir2021MPRNet}, and apply them on images of the selected video trace from UA-DTRAC. Similarly, we use EfficientDet as our detection model. We compare the mAP on each individual raw and enhanced image, and present the results in Fig.~\ref{fig:accuracy_per_image_with_ie}. 

As can be seen, none of selected image enhancement methods makes low-quality samples easier for the detector. In fact, enhanced images lead to worse detection accuracy on some easy samples. The rationale behind that is general purpose image enhancements are usually designed for human visual perception and trained on manually labelled dataset. In real deployments, however, the cause of hard samples with respect to a particular downstream task can be far more complex. For instance, environmental changes (\eg clouds, glare) could result in the drastic lighting condition change within a few seconds. Other common factors include object movements and changes of object sizes. For example, a car gets harder to be detected when it suddenly accelerates or moves away from the camera. In summary, reconstructed details from a single or general purpose image enhancement methods are not sufficiently discriminative for the heavyweight DNN model used in a specific VAP. 

\begin{figure*}[t]
	\centering
	\includegraphics[width=0.95\linewidth]{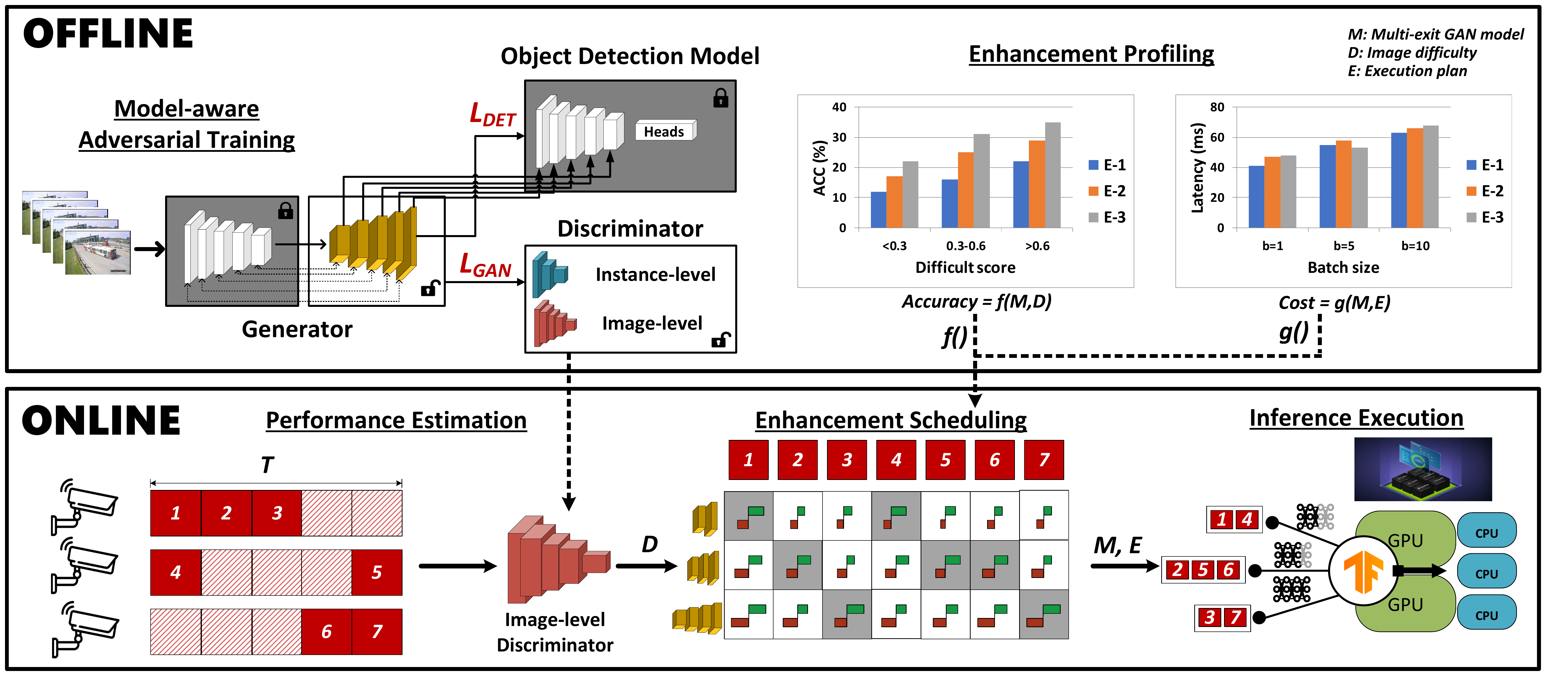} %
	\caption{\name system overview}
	\label{fig:overview}
\end{figure*}

\vspace{5.0em}
\para{Key Takeaways:}
\begin{itemize}
    \item We see a considerable amount of idle GPU compute resource exists in different edge VAPs. However, idle resource availability is highly dynamic and fragmented.

    \item Overall accuracy of a video analytics task can be boosted significantly when the quality of a small portion of frames improves. However, such hard samples are non-uniformly distributed in the time domain and are hard to predict. 

    \item Running off-the-shelf opportunistic enhancement methods, \ie image enhancement in a naive way is inappropriate. It is expensive and could even adversely impact object detection performance. On the contrary, a task-specific enhancement is needed.
\end{itemize}

In what follows, we propose \name, a GAN-based task-specific enhancement model and an online execution scheduler which gracefully select and transform hard frames by harvesting the highly dynamic idle GPU resources.



\section{Design Overview}
\label{sec:sensys_overview}


In \name, We design three key modules, namely discriminator, enhancer, and scheduler. Fig.~\ref{fig:overview} demonstrates the architecture overview of \name, which employs opportunistic enhancement in two phases. In the offline phase, we make attempts to train a discriminator and an enhancer (\ie the Generator in  Fig.~\ref{fig:overview}), which are tailored for the downstream detector. The trained discriminator is thus able to classify if an incoming frame can be well detected. For those hard frames, the trained enhancer provides additional processing which introduces more discriminative details to make the frame more amenable to the detector. \name also provides multiple enhancement levels.
In the online phase, we inject the trained discriminator and enhancer into the VAP, without modifying any other existing modules. The resource-aware scheduler buffers incoming frames and selectively executes the enhancer at different levels within the resource budget, so as to achieving the best overall detection accuracy. 




To train the discriminator and the enhancer jointly in the offline phase, \name's enhancement module builds on top of recent advances in generative adversarial networks (GAN). Unlike traditional general purpose GAN-based image enhancements, \name proposes a task-specific GAN architecture and a model-aware adversarial training mechanism. This GAN aims to improve semantic details for the downstream tasks instead of improving the interpretability or perception of information in images for human viewers. 



To enable fine-grained enhancement at runtime, we devise a multi-exit GAN structure and an adaptive scheduler to decide on-the-fly how the multi-exit GAN is executed on frames so as to maximize the overall object detection accuracy. Due to video content variations and subsampling of the VAP, heavyweight object detection workloads vary over time and would likely under-utilize the GPU resources provisioned upfront for most of the time. Thus, \name's scheduler firstly determines the resource budget by quantifying the number of frames reaching the object detector,
and then uses the discriminator in GAN to classify frames. Based on the classified difficulties and resource budget, a combinatorial optimization problem is formulated to decide at what level the enhancement model is executed on incoming frames.


\section{Adversarial Learning-based Enhancement}
\label{subsec:ie-gan}

GAN~\cite{goodfellow2014generative, zhu2017unpaired, karras2019style,isola2017image} is widely used for the image enhancement and synthesis. GAN adopts the adversarial training~\cite{goodfellow2014generative} to learn a \emph{generator} ($G$) and a \emph{discriminator} ($D$) simultaneously. 

In GAN-based image enhancement, $G$ is responsible for generating synthetic high-quality images, whereas $D$ takes as input both synthetic and real high-quality images, and is trained to distinguish between these two sources.
Since $G$ and $D$ play a competing and continuous game, in which $G$ is learning to produce more and more realistic high-quality images, and $D$ is learning to be better and better at distinguishing synthetic data from real data, GAN-based image enhancement gains both generative and discriminative abilities at the end of training~\cite{jiang2021enlightengan,liang2021swinir,ganIE2021pan,ganIE2019li,ganIE2020du}. 
Such ability well serves the purposes of identifying and transforming hard frame samples to ones that are more amenable to the object detector in the video analytics pipeline.

Despite superior accuracy, GAN is known to be hard to use in reality~\cite{weng2019gan}. The reason is two-fold. First, \emph{general purpose}, \emph{large capacity} GAN model training is challenging and prone to mode collapse, non-convergence and instability due to inappropriate design of network architecture, use of objective function and optimization algorithms~\cite{Zamir2021MPRNet,zeroDCE2020guo,qin2020ffa,liang2021swinir,jiang2021enlightengan}. 
Second, prohibitively high inference cost hinders real-world deployment of GAN. 
Specifically, GAN's generator typically uses a encoder-decoder architecture where latent features are firstly extracted by encoder and then processed by stacked up-sampling layers in decoder for semantic detail recovery. Running $G$ in a naive way on frames in a VAP could incur the high latency. 

\begin{figure}[!t]
	\centering
	\includegraphics[width=0.9\linewidth]{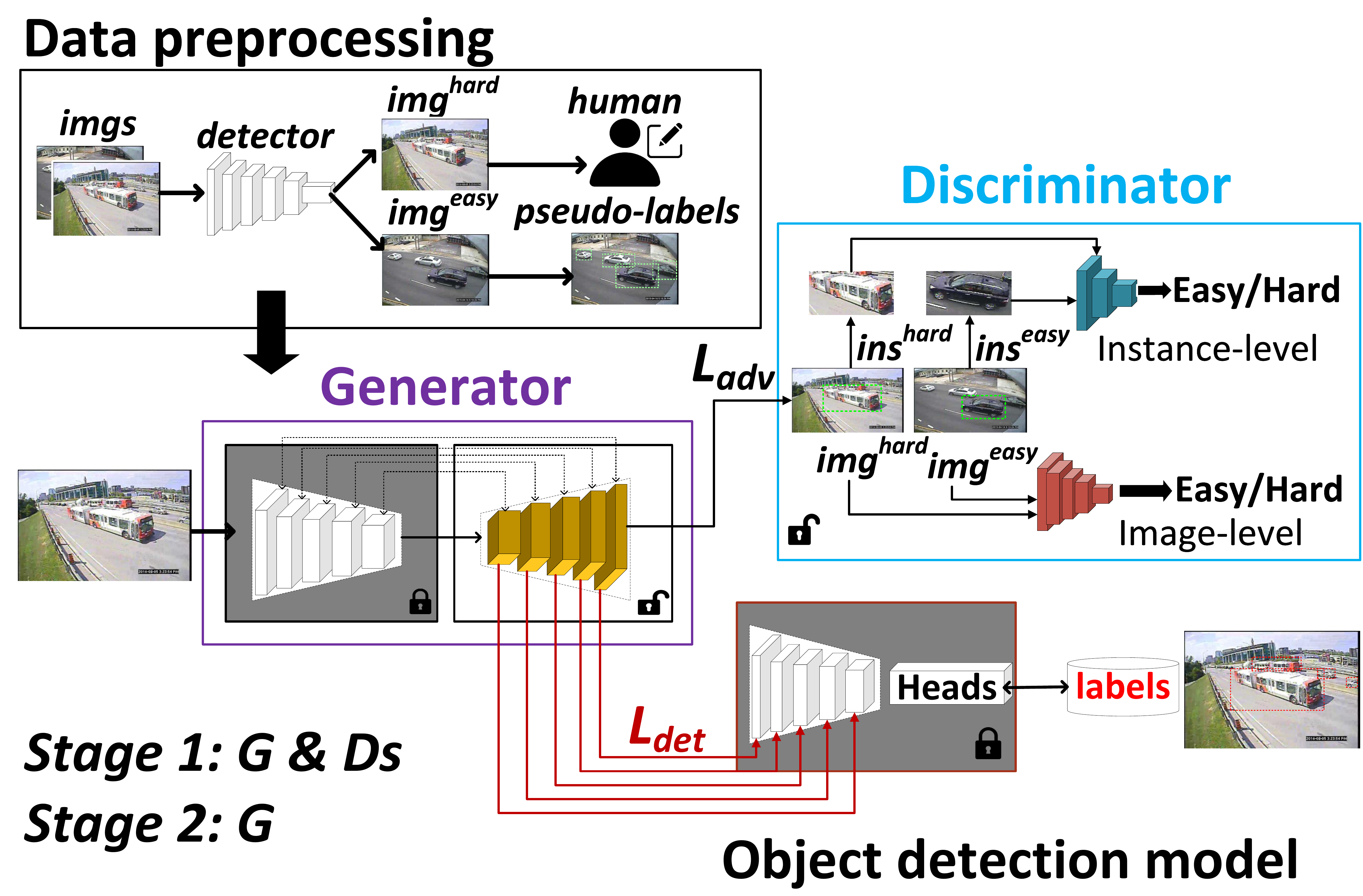}
	\caption{Two-Stage training strategy in \name. In the stage one, we train the generator and discriminators. In the stage two, we fine-tune the generator with the multi-exit mechanism.}
	\label{fig:overview_gan}
\end{figure}

To accelerate the inference pipeline and enable fast adaptation on new testing data, we propose a \emph{detector-specific} GAN architecture and a \emph{model-aware} adversarial training mechanism (Fig.~\ref{fig:overview_gan}). As opposed to building a general purpose model, our GAN architecture is tailored for the object detection model used in a video analytics pipeline. As such, we effectively reduce the complexity of GAN training by learning a general $G$ on a similar public dataset and fine-tuning its discriminator only on testing data. To further reduce training complexity and cost, we replace the encoder in $G$ with backbone layers of the detector since the latter has already been trained for extracting feature embedding. In addition, we introduce a new multi-exit structure between two models which provides more flexibility and fine-grained trade-off between inference latency and accuracy. 



\subsection{GAN Architecture}
\label{subsec:gan_arch}

We follow the common practice~\cite{goodfellow2014generative, zhu2017unpaired, karras2019style,isola2017image, jiang2021enlightengan} to design the overall architecture of GAN. However, two key changes are introduced to our model to make it more suitable for opportunistic enhancement in video analytics.

First, inspired by ~\cite{jiang2021enlightengan}, we design our GAN consisting of one single $G$ and two $D$s, a frame-level $D_f$ and an instance-level $D_i$. $D_f$ is applied on the whole frame to examine if it is hard or not, whereas $D_i$ is applied onto each individual object instance in one frame. The insight behind this design is that one frame could contain hard instances of various types, thus need to be treated differently from each other. 
For the architecture of $D_i$ and $D_f$, we use two and one convolutional layers with three fully connected layers for $D_i$ and $D_f$, respectively.

Second, we reuse the backbone of the downstream DNN to design our $G$. We design the $G$ as a U-Net~\cite{2015unet} architecture, which contains a encoder and a decoder. For a hard frame, the encoder firstly extracts the feature maps, after which the decoder synthesizes an easier frame with more discriminative features from that. Intuitively, the encoder plays an exact role of the backbone network in a detector. Hence, we directly replace the encoder with detector's backbone including its weights. Such a design brings two benefits, compared to training $G$ from scratch:
\begin{inparaenum}[1)]
    \item Since the extracted feature maps are exactly from the backbone, the enhancements based on those would be more specific to the following detector.
    \item Obviously training the decoder only would speedup our training process.
\end{inparaenum}
For the decoder, its architecture should be design according to the backbone's, to form an U-Net.

\subsection{Two-stage Model Training}

We propose a two-stage training process, as illustrated in Fig.~\ref{fig:overview_gan}. In stage one, we try to train a GAN, and in the stage two, we empower the GAN with the multi-exit capability to fit the dynamic idle compute resources. We begin with the stage one.

To train the GAN, a dataset containing training frames is required. We would discuss the training dataset selection in \S\ref{subsec:pre_training}. Here given a set of frames, we firstly identify easy and hard samples using the downstream object detector. Based on the observations that DNNs always perform uncertainly on hard samples, we leverage detector's predicted confidence score as an indicator. Specifically, for frame $F$, we calculate its difficulty score $\theta_F$ by averaging the confidence scores of all its Region-of-Interests (RoIs),
\begin{equation}
\label{eq:difficulty_score}
    \theta_F = \frac{1}{N} \sum^{N}_{i=0} \sigma_i,
\end{equation}
where $\sigma_i$ is the confident score of the $i_{th}$ RoI in the frame $F$.

We select frames with $\theta_F$ lower than a threshold, empirically set to $0.6$, as hard samples, and the remaining as easy ones. By using the selected hard and easy samples, we train the frame-level discriminator $D_f$ and the instance-level discriminator $D_i$. We update their weights via the back-propagation from the following loss functions,
\begin{equation}
    \begin{split}
    L_f & = \mathbb{E}_{x\sim p_{e}(x)}[log D_{f}(x)],\\
    L_i & = \mathbb{E}_{x\sim p_{e}(x)}[log D_{i}(x)],
    \end{split}
\end{equation}
where $L_f$ and $L_i$ are the loss functions for $D_f$ and $D_i$, respectively. $p_{e}$ denotes the data distribution mixed by real and synthetic easy frames. Note that we do not need the annotations of frames to train the discriminators.

Once $D_f$ and $D_i$ are trained, we use them to train the generator $G$. The goal is to make $G$ take as the input a hard frame, and output a \emph{synthetic} easy frame, which $D$ is not able to distinguish from the real easy ones. In particular, we make use of the following adversarial loss in the stage one,
\begin{equation}
    \begin{split}
        L_{S1} &= L_f +  \mathbb{E}_{z\sim p_{h}(z)}[log(1-D_{f}(G(z))] \\ 
        & + L_i + \mathbb{E}_{z\sim p_{h}(z)}[log(1-D_{i}(G(z))]
    \end{split}
    \label{eq:IELoss}
\end{equation}
where $p_{h}$ represents the distribution of hard frame. We then run the adversarial training process, update $G$ and $D$s in the alternating and competing manner, until reaching the convergence.

\subsection{Multi-exit Structure}
\label{subsec:multi-exit}

After the GAN is trained, we connect the output of the decoder with the detector in stage two. Next, we fine-tune the GAN using the annotations of frames without modifying the detector. More specifically, we go through all the frames. For each frame, we run the generator followed by the detector. According to the detection results and the frame annotations (ground truth), we calculate the detection loss $L_{d}$. Based on that we fine-tune and update the decoder of the GAN.

As discussed in \S\ref{sec:sensys_overview}, we propose to enable the elastic execution for the enhancement module, to best utilize the dynamic idle resources. To this end, we introduce a multi-exit structure. Such a design is due to two rationales:
\begin{inparaenum}[1)]
    \item By skipping a different number of layers, the enhancement module would result in multiple accuracy-latency profiles, which are useful for a better scheduling. 
    \item Even for hard frames, they have various difficulty scores. Some of them might be relatively amenable to the detector, a lightweight enhancement is totally enough for a better detection.  
\end{inparaenum}

Specifically, we add connections between the feature maps of the $\kappa_{th}$ layer of the decoder and the $(\beta-\kappa)_{th}$ layer of the detector's backbone, as shown in Fig.~\ref{fig:overview_gan}, where $\beta$ is the total layer number of the backbone. The connections naturally work because we reuse the backbone as the encoder and design the decoder accordingly. As a result, the shape of the connected layers is the same. Furthermore, in our design, not only $(\beta-\kappa)$ layers in the decoder can be skipped, $\kappa$ layers in the detector's backbone can be skipped as well.

Next, we fine-tune the GAN with the multi-exit structure. As we expect all exits to be effective for the final detection results, we compute the detection loss for each exit by comparing its predictions and the annotations, and then average the detection losses of all exits. In this stage, since $G$ is updated, we add $L_{S1}$ to the training process as an regularization term. In sum, the overall loss of the stage two $L_{S2}$ is formulated as,
\begin{equation}
    L_{S2}=\frac{1}{\beta}\sum^{B}_{\kappa=0} L^{\kappa}_{d}+L_{S1}
\end{equation}
where $\kappa$ denotes the exit's index, $L_{d}$ is a standard loss of the object detection, and $\beta$ is the layer number of the backbone.

The multi-exit structure offers several enhancement options with different levels of discriminative characteristics, processing latency and accuracy gain. In \S\ref{sec:scheduler}, we propose an adaptive scheduler to take advantage of this feature.

\subsection{Pre-training Dataset Selection}
\label{subsec:pre_training}

In the offline phase of \name, we pre-train the GAN-based enhancement module, which requires the labelled training data. Intuitively, the performance of \name is highly related to the dataset selection.

A straightforward way to build the training dataset is to collect historical frames from the target scenes, and label them either manually or by a golden model. This approach, however, falls short on efficiency in real deployments. Collecting and labeling a number of frames require mass of human efforts, which could largely impair the practicality.

To this end, we introduce an alternative dataset selection strategy. We propose to use public datasets that contains rich annotations on scenes similar to the target environment. For example, if the deployment focuses on traffic analytics (\eg vehicle counting), the pre-train can be done using BDD100K~\cite{yu2020bdd100k}, which contains more than 100 million annotated frames captured from driving cars. This strategy turned out to work extremely well as in contrast to limited types of hard samples in the target scene, the pre-training process benefits way more from the much larger variety of samples in massive public datasets. In the evaluation (\S\ref{subsec:overall_performance}), we would show more details about the accuracy gains of the dataset selection strategy.
\section{Adaptive Enhancement Scheduling}
\label{sec:scheduler}

To take full advantage of idle resources, we design the adaptive scheduler, which makes the online decisions of applying the most suitable enhancement levels for the incoming frames, maximizing overall detection accuracy within the resource availability. To achieve this, we firstly profile the enhancement module in terms of the latency cost and the accuracy gain, in the offline phase. Then the profilings as well as the pre-trained GAN are deployed on the target edge device. Based on the profilings, in the online phase, we conduct the scheduling. We formulate the scheduling as an optimization problem, and we introduce the heuristic solution to solve it.

\subsection{Enhancement Profiling}
\label{subsec:ie-profiling}

Thanks to the multi-exit design (\S\ref{subsec:multi-exit}), \name provides multiple enhancement levels. Such the diversity makes more rooms for the enhancement scheduling. In order to determine the best enhancement level, we need to get a sense of the latency-accuracy trade-off of executing the pre-trained GAN at different exits.

Profiling the latency of the multi-exit GAN is simple. we run different levels of the enhancement module using various batch sizes in an exhaustive manner. We execute 100 runs and make use of the averaged latency. Finally We obtain $I_\kappa$, which stands for the expected inference latency when executing the $\kappa_{th}$ level enhancement,
\begin{equation}
    I_\kappa = \mu_{D} + \varepsilon_{\kappa} + \nu_{\kappa},
\end{equation}
where $\mu_{D}$ is the inference latency of the frame-level discriminator $D_f$ (\S\ref{subsec:gan_arch}), which is a constant. $\varepsilon_{\kappa}$ donates the latency of the generator $G$ (\S\ref{subsec:gan_arch}) when $\kappa$ layers are used. $\nu_{\kappa}$ donates the latency of the downstream detector when $\kappa$ layers are skipped.

Since we can execute the enhancement with the same level at batch, thus we notate $I_\kappa^n$ where $n$ is the batch size. Note that $\kappa=0$ represents no enhancement would be applied, and $I_0=\mu_{D}+\nu_0$  .

To profile the accuracy gains of the the multi-exit GAN, we bucketize the frames from the training dataset based on their difficulty scores $\theta$ according to Equation~\ref{eq:difficulty_score}. We set the bucket granularity to $0.1$. Then we execute the base detector on every frame, obtain the accuracy without the enhancement, \ie mAP. Next we execute each enhancement level of the GAN on every frame, and obtain the corresponding mAP improvement compared to the base detector. Finally we average the mAP of frames in the same bucket, and obtain $P_\kappa^\theta$, which stands for the expected accuracy gains when applying the $\kappa_{th}$ level enhancement on a frame with the difficulty of $\theta$. Note that $\kappa=0$ represents no enhancement would be applied, and $P_0=0$.

Note that the profiling is a one-time effort and we put in the offline phase.


\subsection{Heuristic-based Optimization}
\label{subsec:scheduler-formulation}

With the latency profile $I_\kappa^n$ and the accuracy profile $P_\kappa^\theta$, we formulate the scheduling as an optimization problem. Given $M$ frames streamed from multiple cameras, they are required to be processed within $T$, which is the latency constraint of the VAP. Due to the filtering modules, only $m$ frames need to be processed where $m<M$. The scheduling purpose is to generate an enhancement plan. In particular, for each frame $x, x \in m$, we determine a $\kappa$, to maximum the total accuracy gains of all the $m$ frames, while the total latency is not beyond the constraint $T$. Specifically we formulate it as,

\begin{align}
    \max \quad & \sum P_{\kappa}^{\theta_x^{\prime}}, x \in m, \\
	s.t. \quad & f(\sum I_\kappa) \le T,
\end{align}
where $\theta_x^{\prime}$ is the estimated difficulty score of the frame $x$, using the frame-level discriminator $D_f$. The function $f(\cdot)$ is to organize the frames assigned by the same enhancement level to execute in a batch.

Suppose there are $m$ frames and $\beta$ enhancement levels in total. Then the search space of the optimization problem would be $\beta^m$. Particularly, this problem is one kind of non-linear generalized assignment problem (GAP)~\cite{DAMBROSIO2020104933}. It is known to be NP-hard. Though the search space $\kappa$ might be limited in real deployments (4-5), it still brings the scheduling overhead. To this end, in \name we devise sub-optimal solution with two heuristics:
 \begin{inparaenum}[1)]
    \item the pre-trained multi-exit GAN has a monotonic characteristic that the higher enhancement level is involved, the more accuracy improvement is achieved (see \S\ref{subsec:multi-exit-eval}). 
    \item Applying enhancement on the hardest frames tends to yield higher marginal accuracy gains.
 \end{inparaenum}
 
 As such, we follow a prune-and-search approach in three steps. 
 \begin{inparaenum}[1)]
    \item We assign all the $m$ frames with the maximum $\kappa$.
    \item Since this enhancement plan would most likely violate latency constraint $T$, so we select the frame that has the minimal marginal accuracy gain, and assign $\kappa-1$.
    \item We repeat the prior step until the $T$ is met.
 \end{inparaenum}
 Once the enhancement plan is determined, we execute each frame according to the plan.

\section{Evaluation}
\label{sec:evaluation}

\begin{figure*}[!ht]
    \begin{minipage}{0.32\linewidth}
        \centering
    	\includegraphics[width=\linewidth]{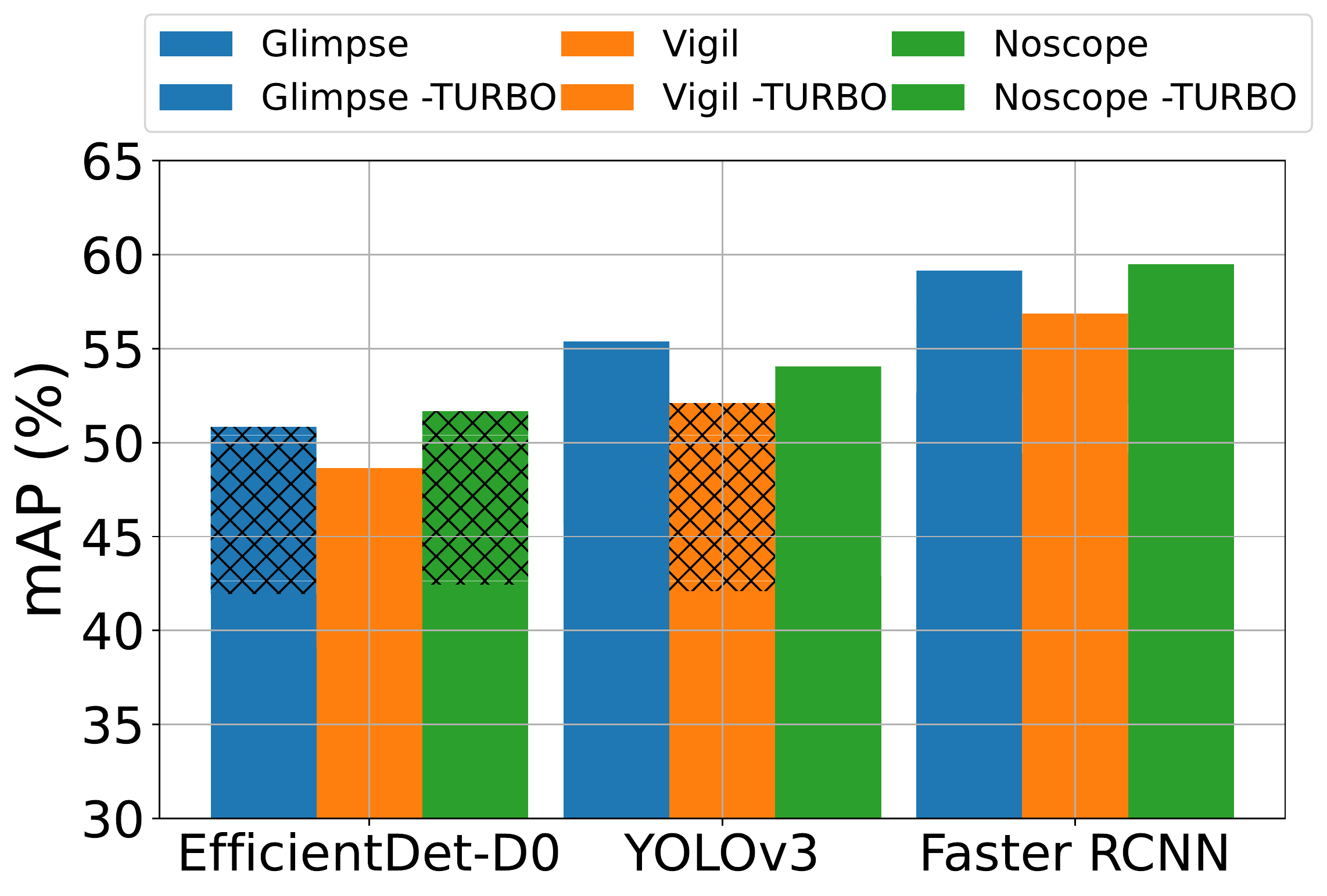}
    	\caption{Overall mAP on UA-DETRAC of the selected VAPs and the enhanced VAPs by \name, with different object detectors, on T4 GPU.}
    	\label{fig:overall_detrac_t4}
    \end{minipage}
    \hfill
    \begin{minipage}{0.32\linewidth}
        \centering
    	\includegraphics[width=1.0\linewidth]{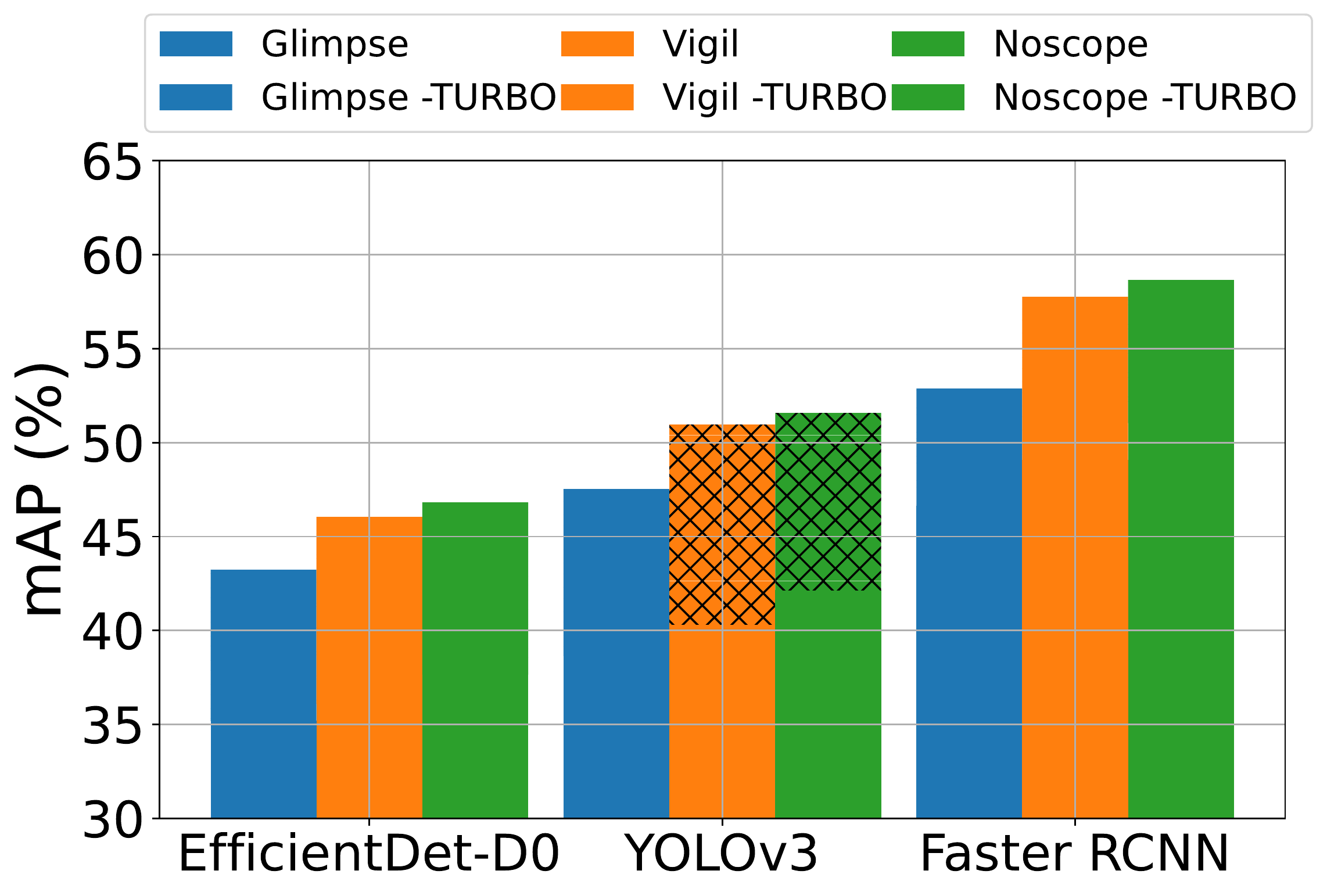}
    	\caption{Overall mAP on AICity of the selected VAPs and the enhanced VAPs by \name, with different object detectors, on T4 GPU.}
	\label{fig:overall_aicity_t4}
    	\label{fig:overall_ai_t4}
    \end{minipage}
    \hfill
    \begin{minipage}{0.32\linewidth}
    	\centering
    	\includegraphics[width=1.0\linewidth]{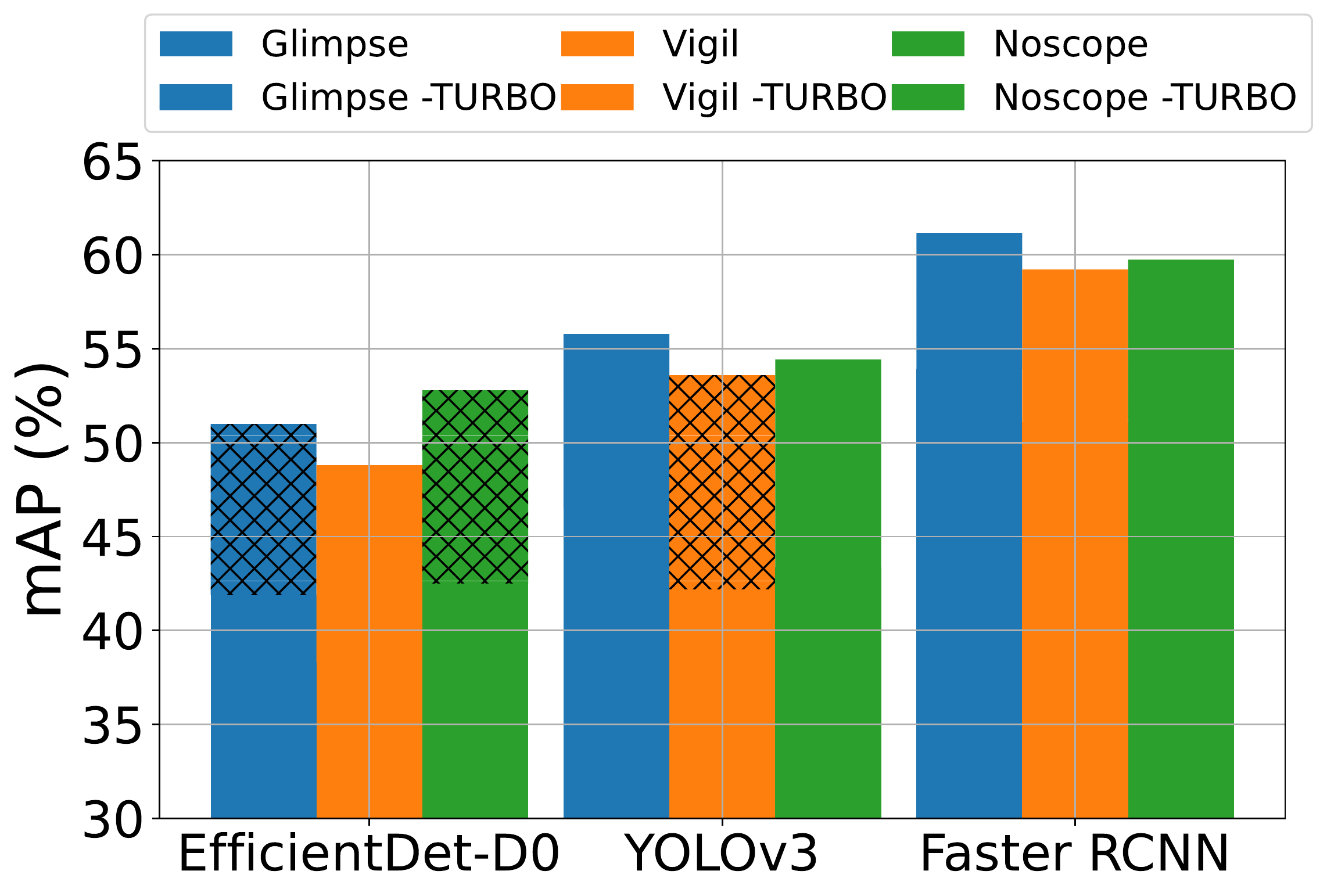}
    	\caption{Overall mAP on UA-DETRAC of the selected VAPs and the enhanced VAPs by \name, with different object detectors, on V100 GPU.}
    	\label{fig:overall_detrac_V100}
    \end{minipage}
\end{figure*}

In this section, we present evaluation results of \name with three canonical VAPs on two real-world video datesets.


\subsection{Experimental Setup}
\label{subsec:exp-setup}

\textbf{Video Analytics Pipeline (VAP)}. Various kinds of video analytics pipelines have been developed to strike the balance between inference accuracy and compute/network cost~\cite{measurement2021xiao,jiang2018chameleon,zhang2017live,Reducto,glimpse2015,vigil2015,noscope2017,ga19mobisys}. In our experiments, we adopt three canonical cascaded VAPs with the object detection as the downstream task. 
\begin{enumerate}
    \item \textbf{Glimpse}~\cite{glimpse2015} measures inter-frame difference, and sends only frames that contain new objects to the object detector down the pipeline. 
    \item \textbf{Vigil}~\cite{vigil2015} runs a lightweight object detection model firstly, and sends only frames that contain most objects to the edge cluster for the heavyweight DNN inference. 
    \item \textbf{NoScope}~\cite{noscope2017} identifies frames with significant pixel changes, and runs a cheap DNN to select frames with low confidence, then calls the edge-based heavyweight object detection. 
\end{enumerate}

\textbf{Dataset.} We evaluate \name on two traffic video datasets, UA-DETRAC~\cite{CVIU_UA-DETRAC} and AICity~\cite{Tang19CityFlow}. The videos are captured by the surveillance cameras on streets. UA-DETRAC consists of 10 hours of videos captured at 24 locations with 25 FPS and the resolution of $960\times 540$. AICity contains more than 3 hours of videos at 10 FPS with resolution of at least $1920\times 1080$. UA-DETRAC includes three types of vehicles annotations, including car, bus and van. AICity consists of two types of vehicles annotations, which are car and van. In total there are $128,800$ frames and $1,103,160$ annotations for our evaluation.

In our experiments, we also use BDD100K~\cite{yu2020bdd100k} to pre-train our GAN enhancement module. BDD100K contains 100K driving videos collected from more than 50K rides. The videos are not from the surveillance cameras, but the targets are vehicles as well. Therefore we use it as the pre-training dataset to verify our designs in \S\ref{subsec:pre_training}.


\textbf{Object Detection Models.} We train and evaluate our GAN enhancement module upon three popular DNN object detectors, including YOLOv3~\cite{yolov3}, Faster RCNN~\cite{fasterrcnn2015} and EfficientDet~\cite{efficientdet2020}. All of models are pre-trained on COCO dataset~\cite{coco2014}.


\textbf{Metrics.} We use the mAP to measure the analytics accuracy of the selected VAPs. We use the streaming multiprocessors (SM) utilization~\cite{nvidia-cuda}, a fine-grained metric for GPU utilization, to quantify the compute resource utilization.

\textbf{Test platforms.} We make use of two platforms, the Azure Stack Edge Pro~\cite{azure_stack_edge} with a NVIDIA Tesla T4 GPU~\cite{nvidia_t4}, and a virtual machine equipped with a NVIDIA Tesla V100 GPU. T4 has 320 Tensor cores and 16GB GPU memory, V100 GPU has 640 Tensor cores and 16GB GPU memory.

\textbf{Implementation.} We build and train the models in \name with TensorFlow~\cite{tensorflow2015-whitepaper}, and execute the inference using NVIDIA Triton Inference server~\cite{triton2019}. GPU SM utilization is collected using NVIDIA DCGM~\cite{dcgm2021}. We implement an application to simulate video streams from multiple cameras over HTTP. In our evaluation, we feed four video streams to an edge node. 


\subsection{End-to-End Evaluation}
\label{subsec:overall_performance}

Fig.~\ref{fig:overall_detrac_t4} illustrates the overall performance of \name on UA-DETRAC with T4 GPU, compared to the selected VAPs integrated with three DNNs.  Overall, \name improves the absolute mAP by $9.02\%$, $11.34\%$ and $7.27\%$ on average for EfficientDet-D0, YOLOv3 and Faster RCNN across the selected VAPS. Specifically, for Vigil integrated with YOLOv3, \name can bring about 9.35\% mAP improvement. Without the enhancement, the Glimpse with Faster RCNN achieves 53.42\% mAP, and \name can bring 5.70\% more mAP to the VAP.

The averaged mAP improvements of the VAPs, Glimpse, Vigil and NoScope across different detectors are $8.32\%$, $9.20\%$ and $8.66\%$, respectively. \name achieves the higher mAP improvements in Vigil and NoScope. It is because that VAPs with model-based pruning are prone to feed more hard frames, so are benefiting from \name much more than temporally filtered frames from Glimpse.

\textbf{Performance over different datasets.} In addition to UA-DETRAC, we also evaluate \name on AICity dataset as well. Shown in Fig.~\ref{fig:overall_aicity_t4}, on the AICity dataset, \name performs even better than on UA-DETRAC. For instance, with and without the \name enhancement, Glimpse integrated with YOLOv3 could obtain 38.71\% and 47.32\% mAP, respectively. It is because that the pre-trained detectors have poor performance on AICity, while \name significantly enhances these imperfect detectors. Overall the averaged mAP improvements of Glimpse, Vigil and NoScope across different detetectors on AICity are $12.46\%$, $9.71\%$ and $8.42\%$, respectively.


\textbf{Performance over different devices.} Different devices might expose different idle resources. To show the effectiveness of \name on other computing devices, we evaluate \name on UA-DETRAC dataset with V100. As shown in Fig.~\ref{fig:overall_detrac_V100}, \name achieves a higher mAP improvement than T4 (Fig.~\ref{fig:overall_detrac_t4}) for three object detectors. Especially on Faster RCNN, \name boosts 4.45\% absolute mAP on average. It is because that idle resources on V100 is much more than T4.    

\begin{figure*}[!ht]
    \begin{minipage}{0.32\linewidth}
        \centering
	    \includegraphics[width=1.0\linewidth]{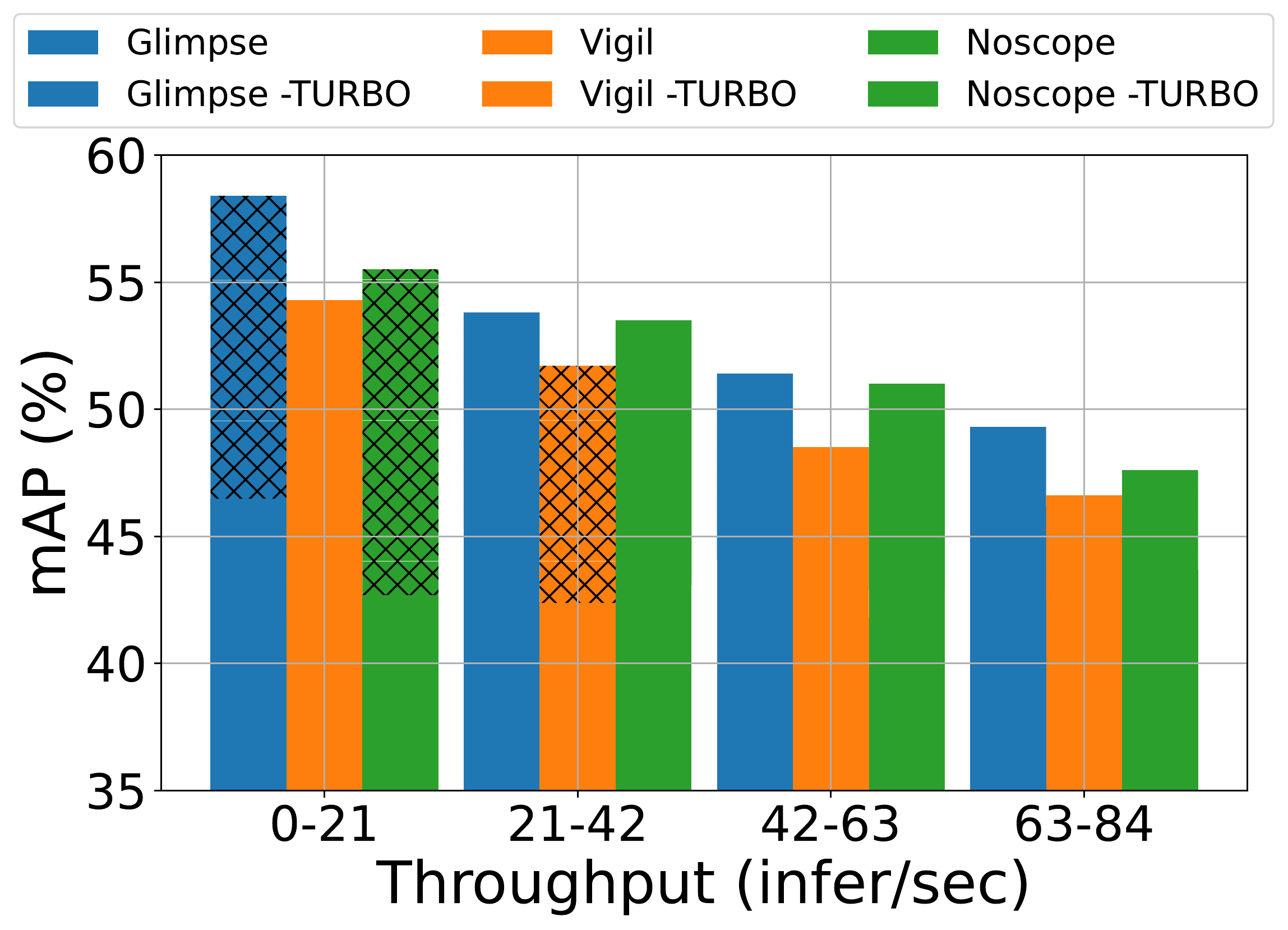}
	    \caption{mAP of the selected VAPs and the enhanced VAPs by \name with YOLOv3 under different throughput, using T4 GPU on UA-DTRAC.}
	\label{fig:overall_detrac_T4_yolov3}
    \end{minipage}
    \hfill
    \begin{minipage}{0.32\linewidth}
        \centering
    	\includegraphics[width=1.0\linewidth]{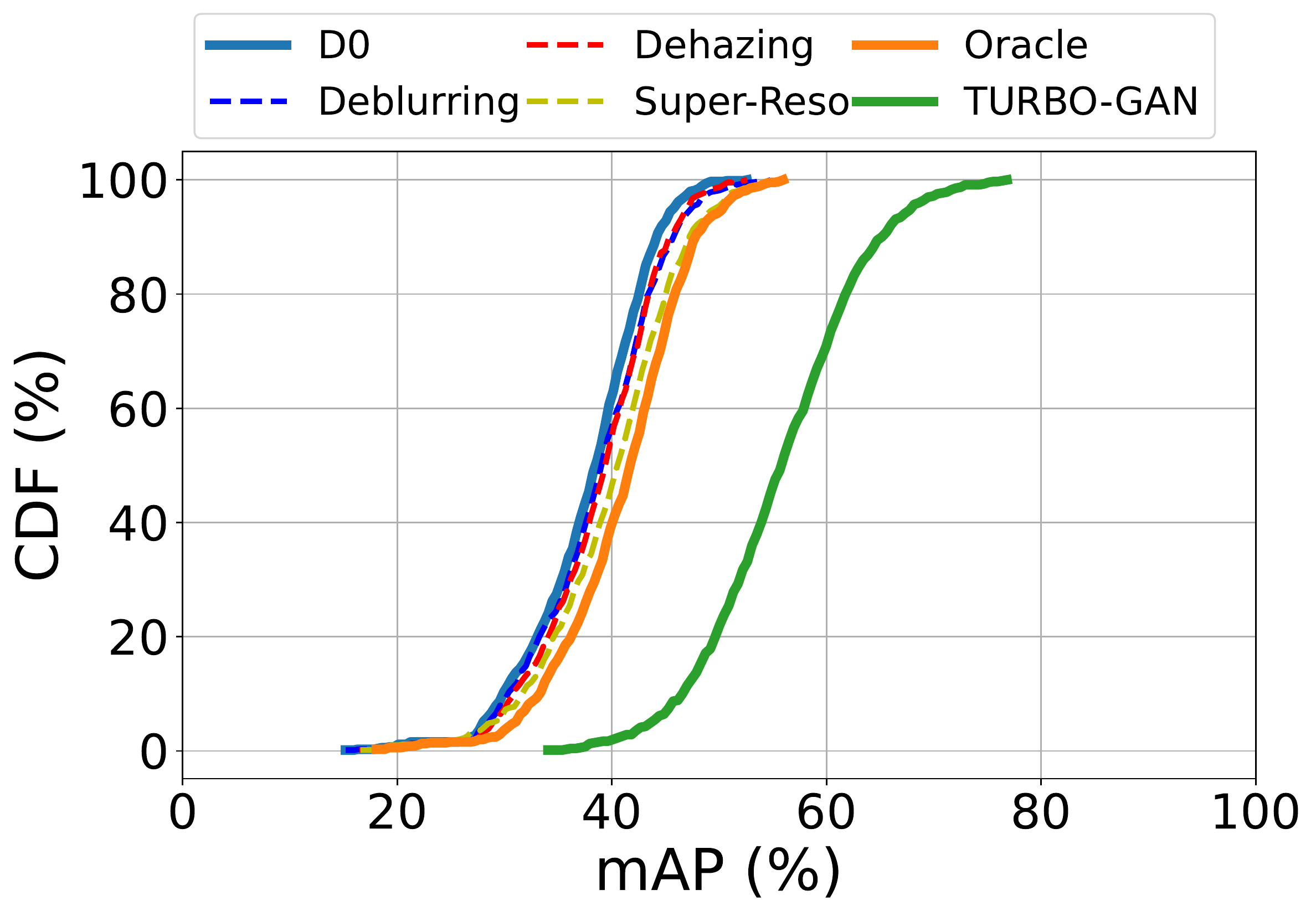}
    	\caption{The mAP distribution of frames in UA-DETRAC obtained by the \name, compared to different image enhancement baselines.}
    	\label{fig:enhance_detrac_glimpse}
    \end{minipage}
    \hfill
    \begin{minipage}{0.32\linewidth}
    	\centering
	    \includegraphics[width=1.0\linewidth]{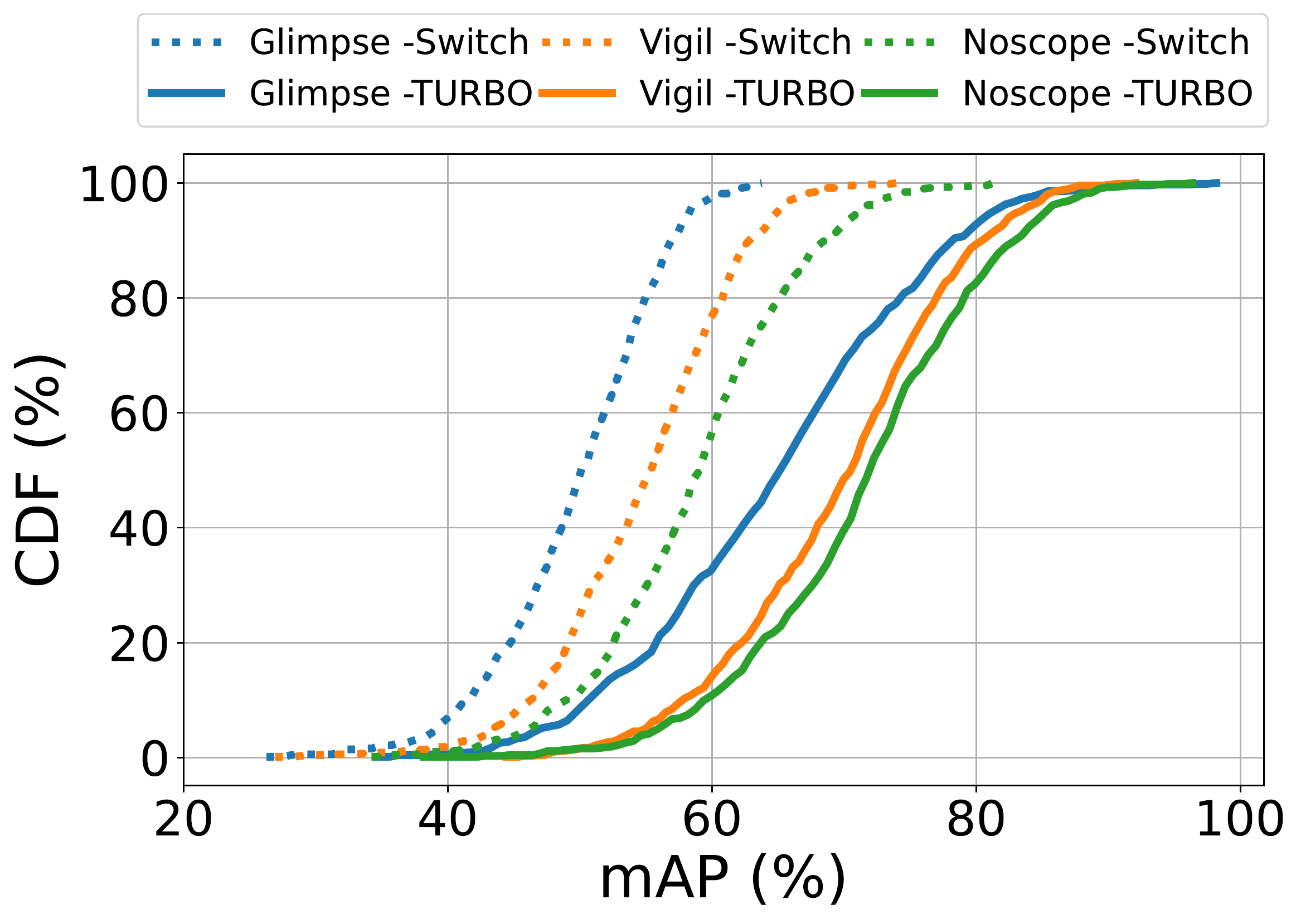}
        \caption{The mAP distribution of frames in UA-DETRAC obtained by the \name, compared to different model switching baselines.}
        \label{fig:mAP_baselines}
    \end{minipage}
\end{figure*}

\textbf{Performance over different throughout.} To understand how \name harvest the idle resources in fine-grained ways, we introduce the performance over different throughout. As described in \S\ref{subsec:edge_vap}, throughput is the real arriving rate of video streams, it represents the number of frames the VAP need to process per second. According to our evaluation, the maximum processing capacity of T4 for YOLOv3 is 84 inference per second (infer/sec). Fig.~\ref{fig:overall_detrac_T4_yolov3} illustrates the obtained mAP of the enhanced VAPs under different throughout. \name perform significantly well when the throughout is low since more idle resources are harvested to enhance the VAP. Particularly, 13.75\% absolute mAP improvement can be obtained for all of VAPs when the throughput is lower than 21 infer/sec.


\textbf{Performance over more baselines.} In addition to the raw VAP, we also introduce two more baselines, which are potentially used as the opportunistic enhancement approaches, the image enhancement and the model switching.

For the baseline of image enhancement, we replace the generator of the GAN with the image enhancement models. We use three popular image enhancement models, including deblurring~\cite{Zamir2021MPRNet}, dehazing~\cite{qin2020ffa} and super resolution~\cite{liang2021swinir}. We also add an oracle image enhancement, which would select the best image enhancement model for each frame. Fig.~\ref{fig:enhance_detrac_glimpse} shows the evaluation results of \name compared to the baseline of image enhancements. Although the oracle enhancement improves $3.34\%$ average mAP, it is still much lower than \name, which gets $13.51\%$ mAP improvement.

For the baseline of model switching, we runs our scheduling algorithm on a model zoo including, EfficientDet-D0, D4 and D7, to use the larger model to harvest the idle resources. Fig.~\ref{fig:mAP_baselines} shows the evaluation results of \name compared to the model switching baselines. \name achieves more than 12.4\% mAP than the best model switching baseline.



\begin{figure*}[!t]
\subfigure[EfficientDet-D0]{
\begin{minipage}{0.32\textwidth}
        \centering
        \includegraphics[width=\textwidth]{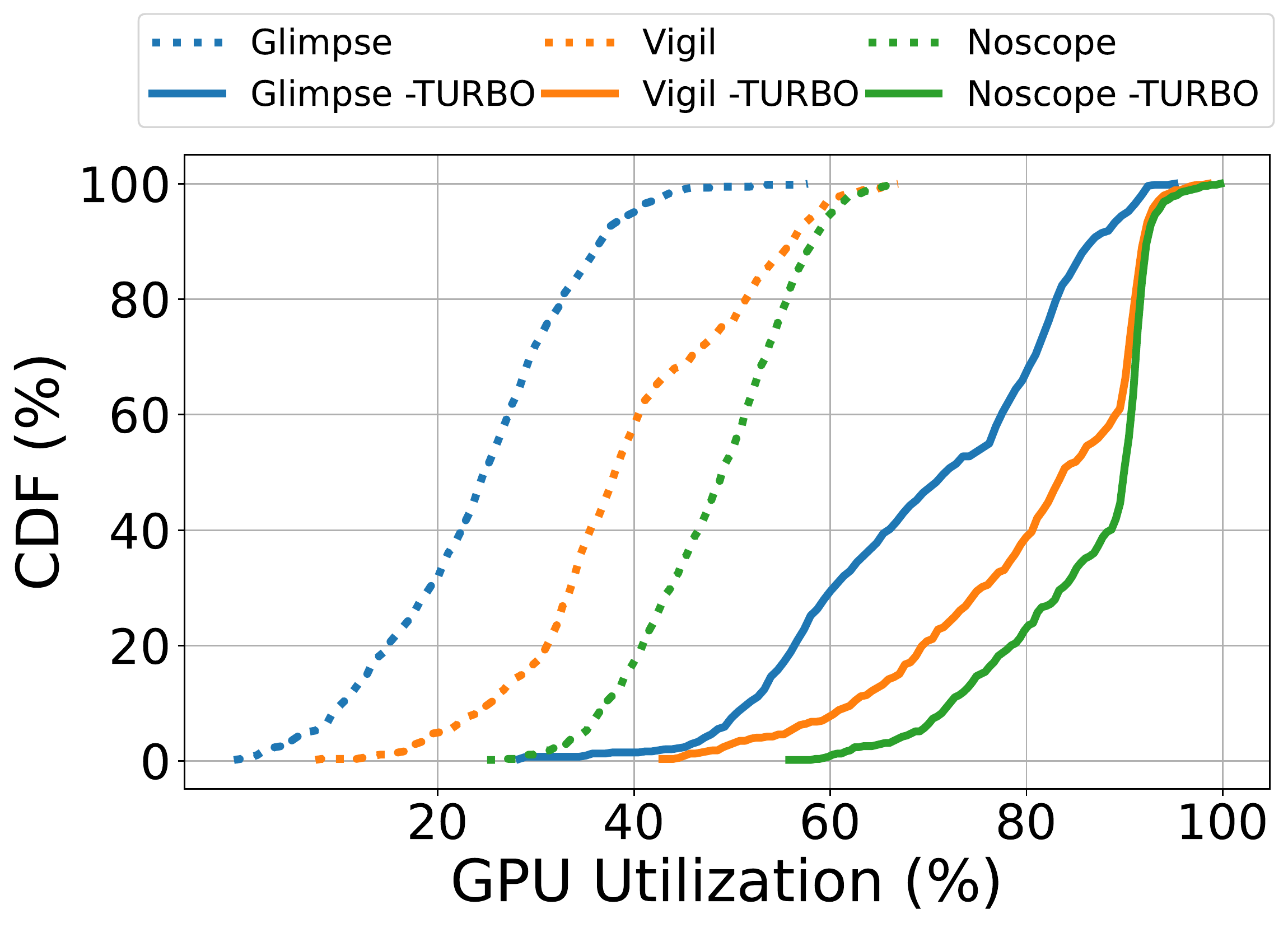}
        \label{fig:overall_map_glimpse}
\end{minipage}}
\hfill
\subfigure[YOLOv3]{
\begin{minipage}{0.32\textwidth}
        \centering
        \includegraphics[width=\textwidth]{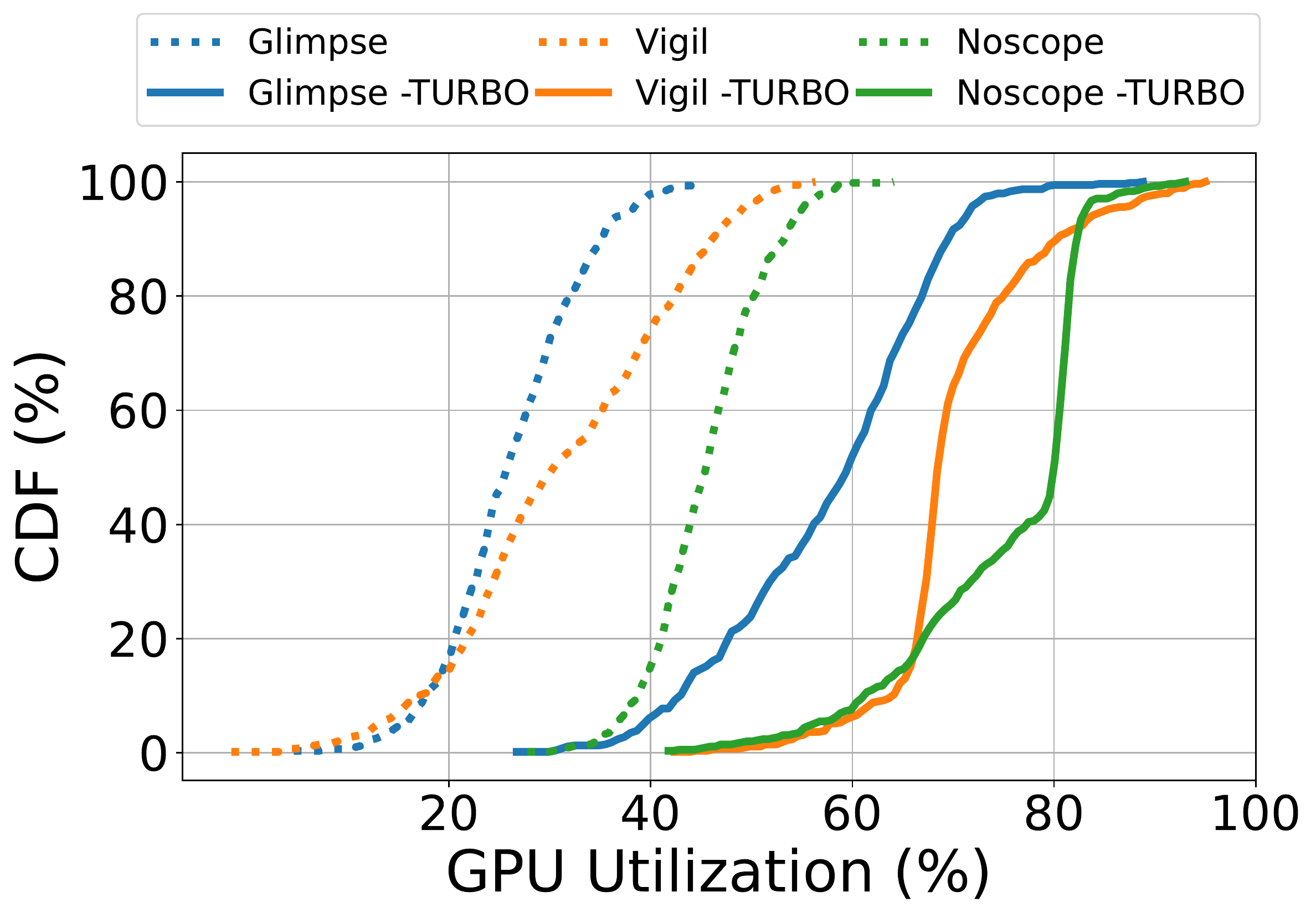}
        \label{fig:rcnn_map_vigil}
\end{minipage}}
\hfill
\subfigure[Faster RCNN]{
\begin{minipage}{0.32\textwidth}
    \centering
    \includegraphics[width=\textwidth]{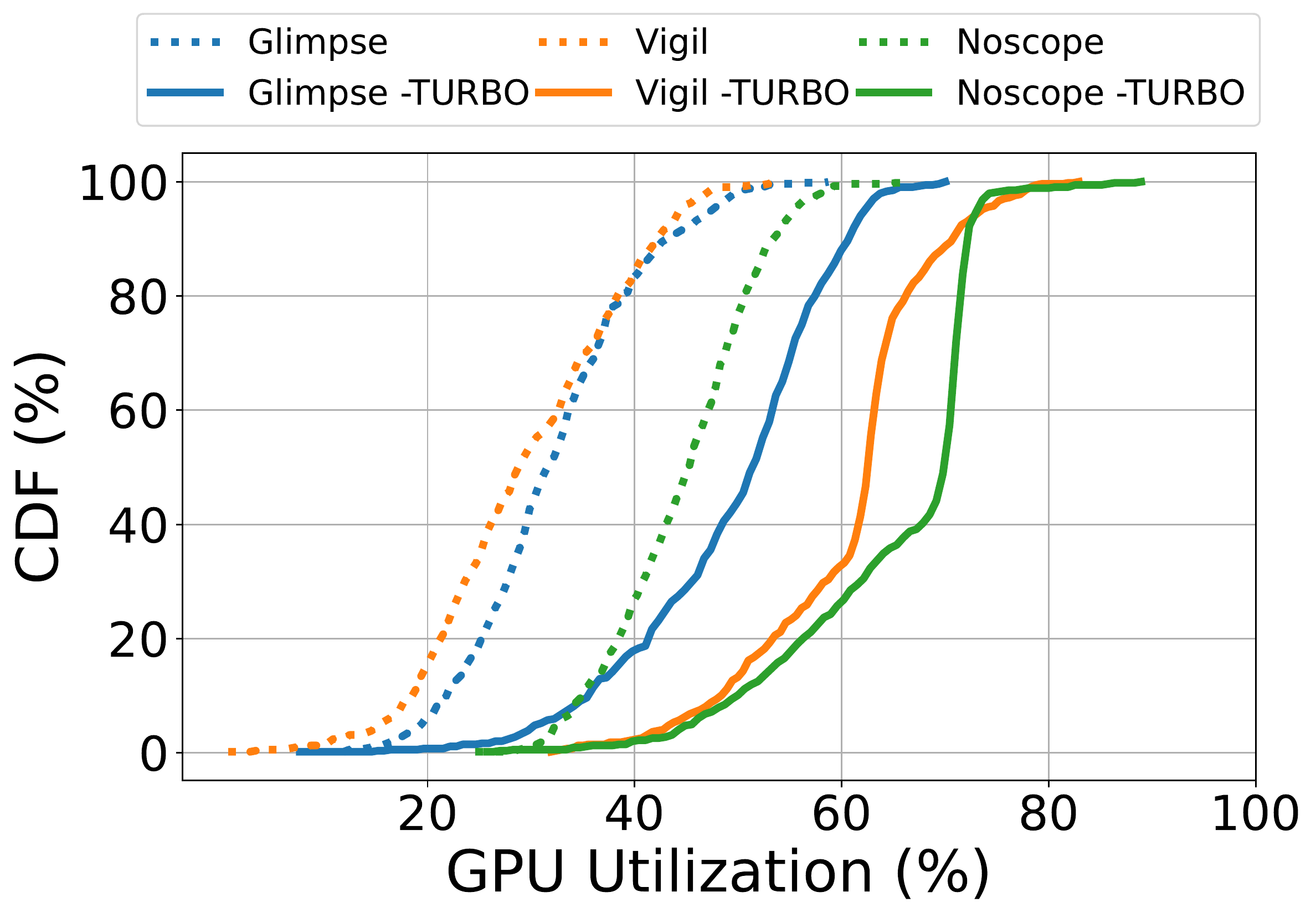}
    \label{fig:rcnn_map_noscope}
\end{minipage}}
\caption{GPU utilization of \name with the selected VAPs integrated with different object detectors on T4.}
\label{fig:rcnn_map_gpu}
\end{figure*}

\textbf{GPU utilization and overhead.} \name harvests the idle GPU resources to enhance the VAP. Therefore we evaluate the GPU utilization. As shown in Fig.~\ref{fig:rcnn_map_gpu}, \name successfully utilizes more idle GPU resources, resulting in the higher GPU utilization rate. Particularly, \name brings 25.42\% more GPU utilization on the T4 GPU.

\name is also lightweight. Specifically, the discriminator cost around 10$ns$ on T4, and the full generator would only cost 23.31$ms$. As the reference, YOLOv3 would cost 35.71$ms$ per inference on the T4 GPU.


\begin{figure}[!t]
    \centering
	\includegraphics[width=0.85\linewidth]{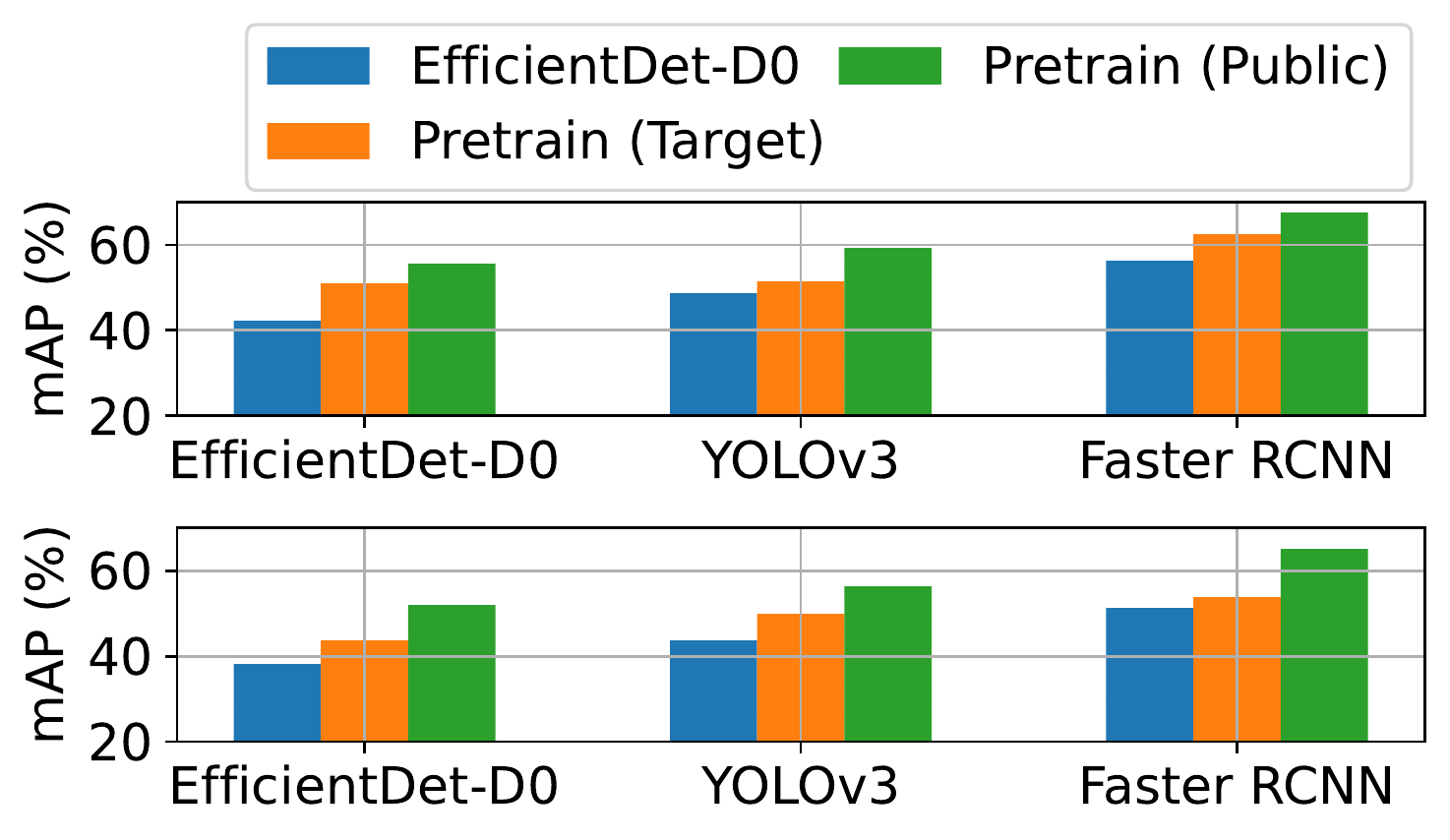}
\caption{Achieved mAP of \name pre-trained on target dataset and Public dataset.}
\label{fig:pretrain}
\end{figure}

\subsection{Evaluation of Multi-Exit GAN}
\label{subsec:multi-exit-eval}
Next we break down \name and evaluate key components in detail.

\textbf{Effectiveness of the pre-training strategy.}
To show effectiveness of our data selection on GAN pre-training, we show mAP of our GAN pre-trained on two datasets, shown in Fig.~\ref{fig:pretrain}. In specific, we train our GAN for EfficientDet-D0 on the target dataset and a public dataset separately. In our experiments, we select BDD100K~\cite{yu2020bdd100k} as the public dataset to train GAN. 

It is interesting to note that GAN pre-training on target datasets are hard to preserve mAP improvements for all detection models. But pre-training GAN on a large-scale public dataset achieves a stable mAP improvement on different models or datasets. It is because that target datasets cannot provide enough hard samples for training \name's GAN. For instance, there are only 960 hard samples on the target scenes (AICity) and 65,540 hard samples on BDD100K for Faster RCNN.

\begin{figure*}[!t]
\subfigure[EfficientDet-D0]{
\begin{minipage}{0.32\textwidth}
        \centering
        \includegraphics[width=\textwidth]{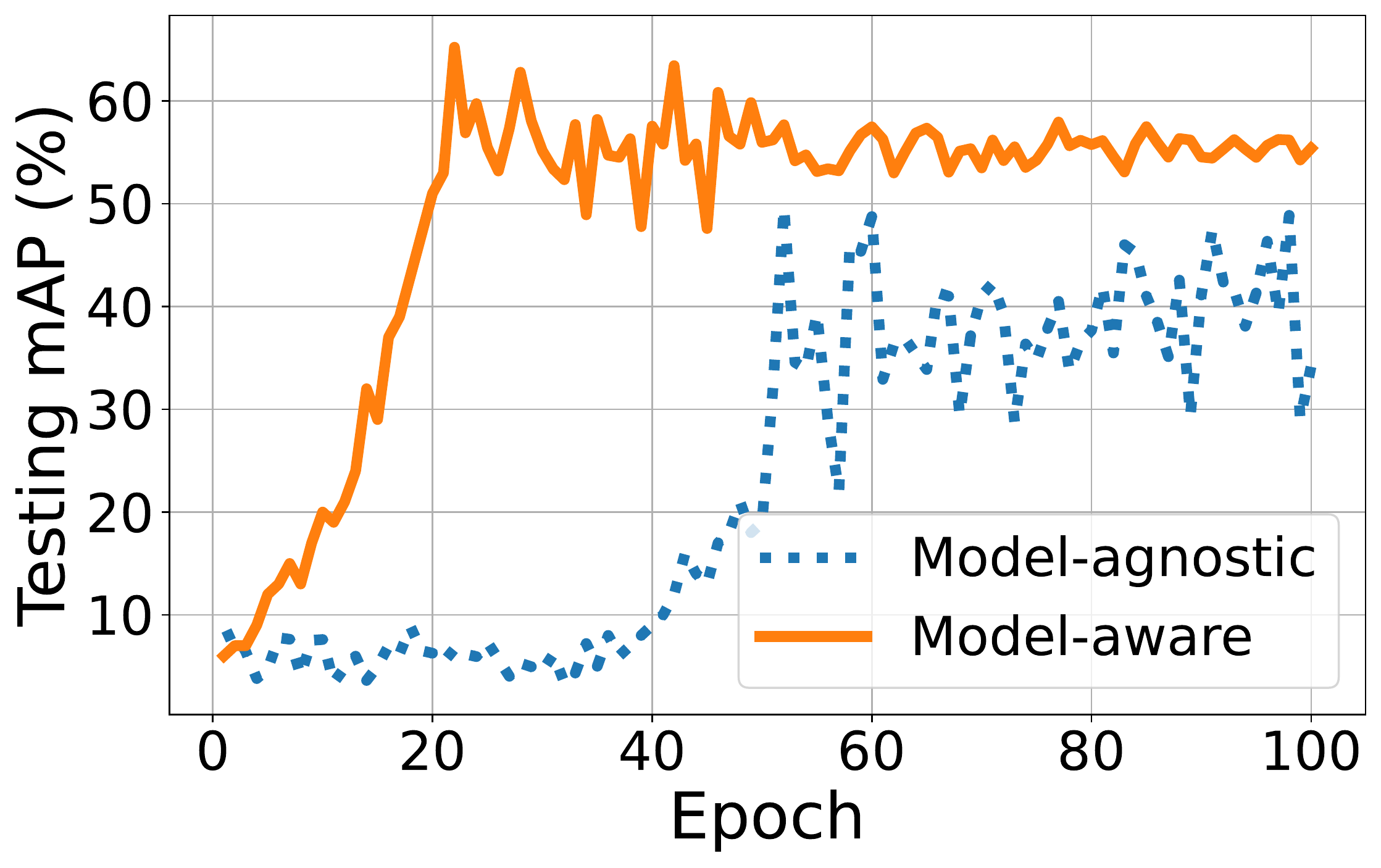}
        \label{fig:efficientdet_specific}
\end{minipage}}
\hfill
\subfigure[YOLOv3]{
\begin{minipage}{0.32\textwidth}
        \centering
        \includegraphics[width=\textwidth]{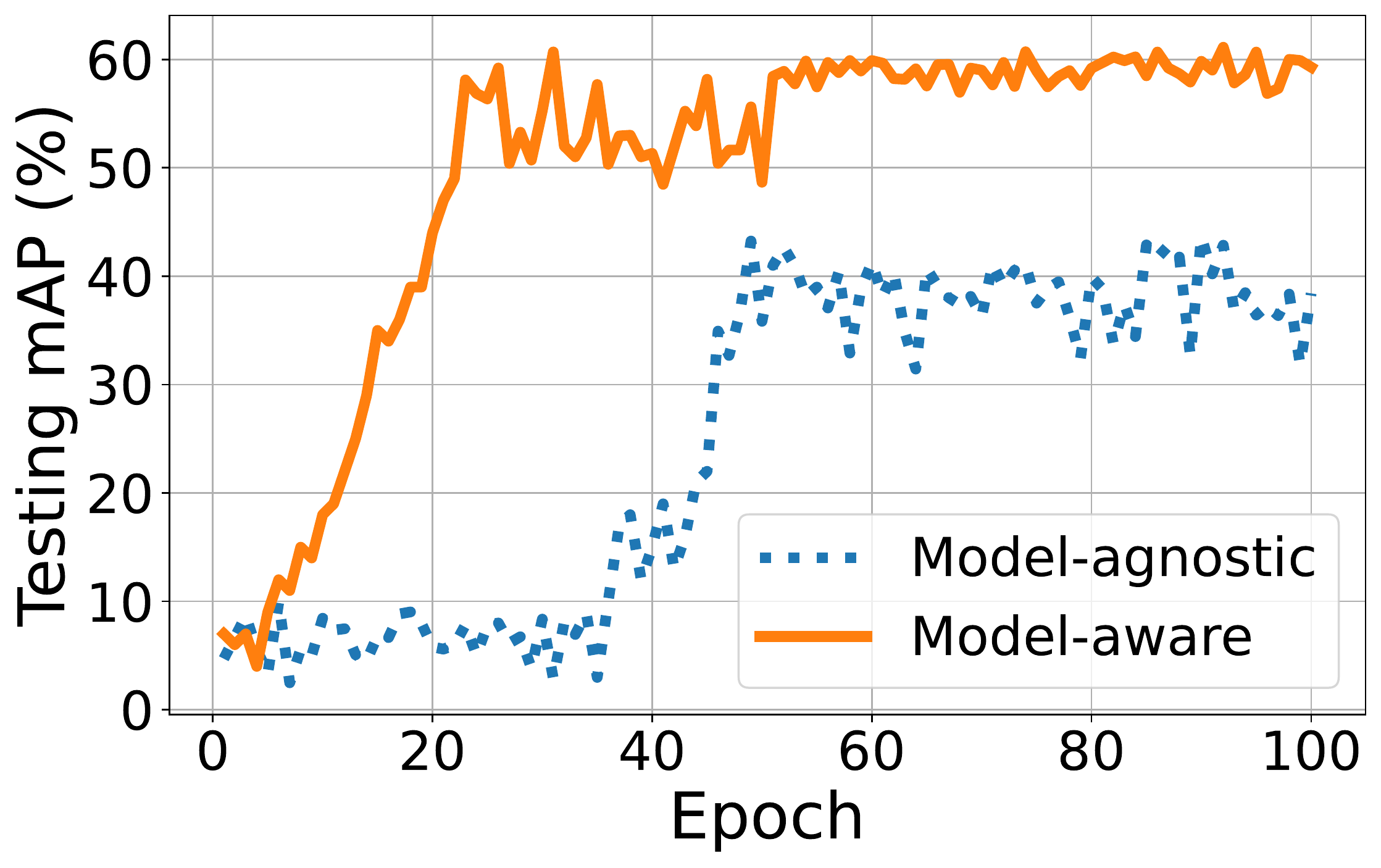}
        \label{fig:yolov3_specific}
\end{minipage}}
\hfill
\subfigure[Faster RCNN]{
\begin{minipage}{0.32\textwidth}
    \centering
    \includegraphics[width=\textwidth]{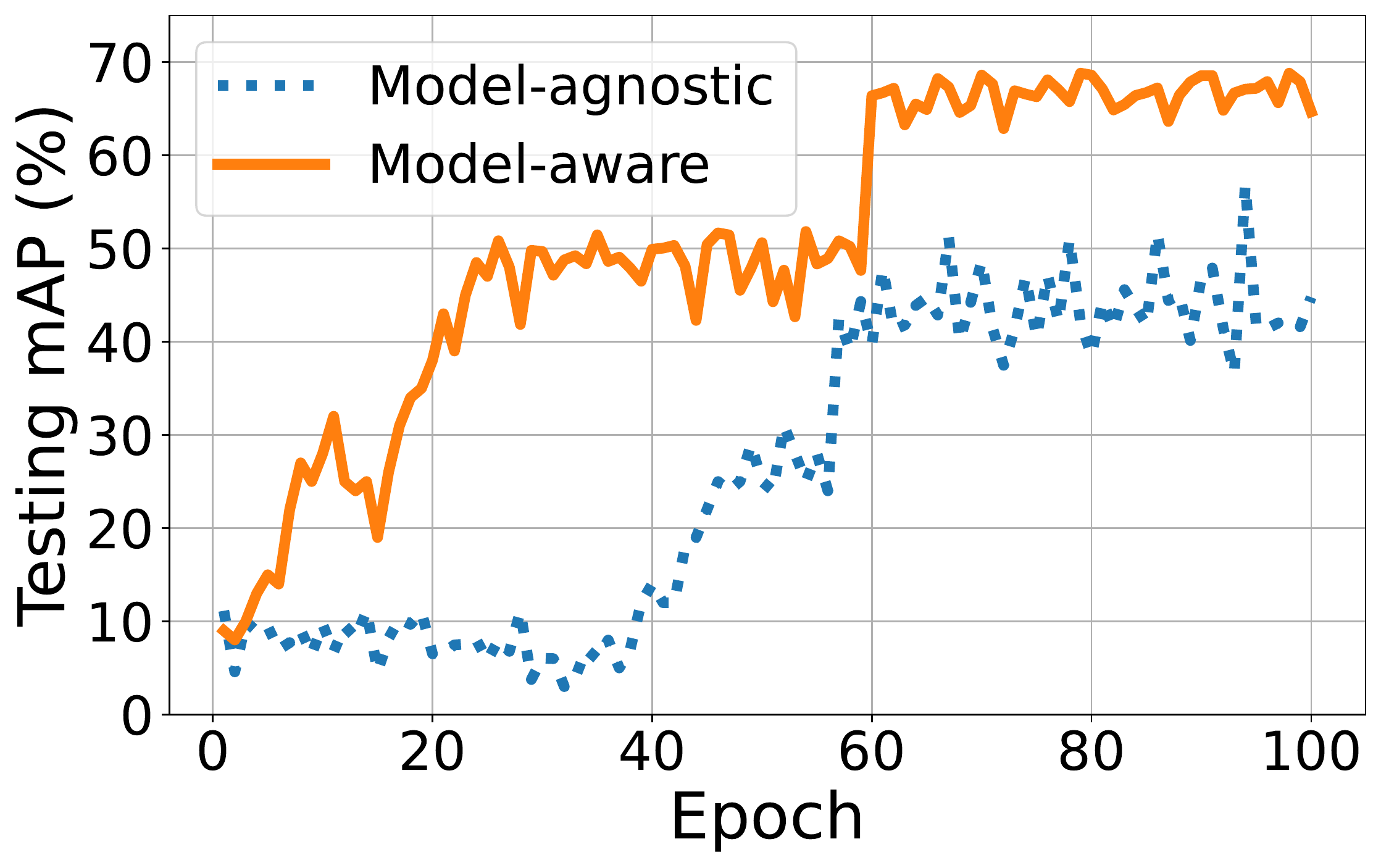}
    \label{fig:rcnn_specific}
\end{minipage}}
\caption{The test mAP of each training epoch using model-aware and model-agnostic training strategies.}
\label{fig:adversarail_training_gpu}
\end{figure*}

\textbf{Effectiveness of the model-aware adversarial training.}
We propose a model-aware adversarial training, therefore we compare with a model-agnostic adversarial training strategy. It selects frames with more small objects as hard samples, since the scale-variant is one of challenges for locating objects. Objects with small scales are hard to be annotated in existing large-scale labeled training datasets (\eg Microsoft COCO~\cite{coco2014}, UA-DETRAC~\cite{CVIU_UA-DETRAC} and MIO-TCD~\cite{luo2018miotcd}). 

We set the same initial model architectures and parameters for both training strategies. Model-agnostic training fine-tunes all layers end-to-end instead of freezing backbone layers. As in Fig.~\ref{fig:adversarail_training_gpu}, model-aware training only requires $1/3$ time of model-agnostic training but achieves $15\%$, $18\%$ and $22\%$ more mAP improvements for EfficientDet-D0, YOLOv3 and Faster RCNN.

\begin{figure*}[!t]
    \centering
    \subfigure[EfficientDet-D0]{
    \includegraphics[width=.3\textwidth]{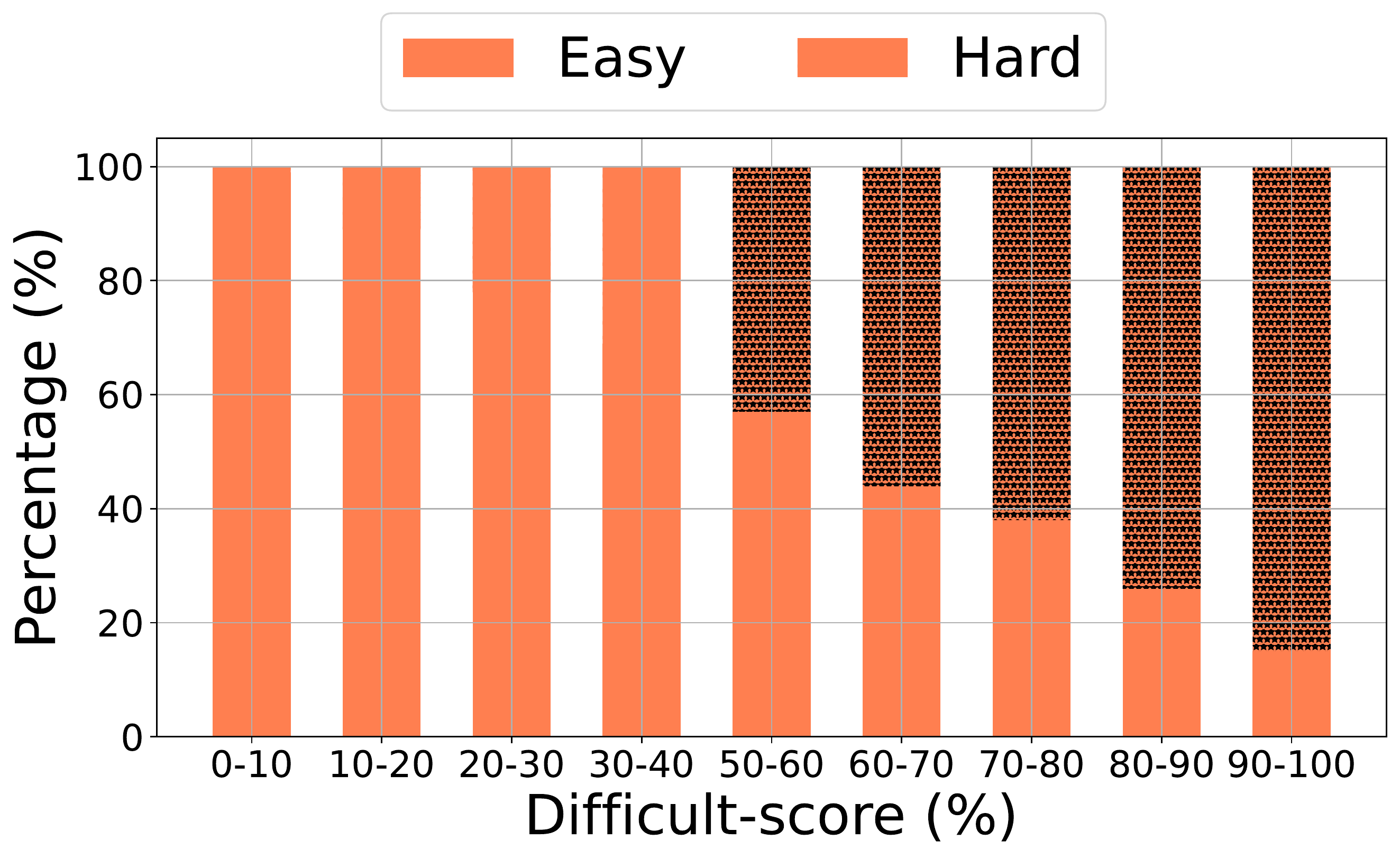}
    \label{fig:discriminator_efficientdet0}
    }
    \quad
    \subfigure[YOLOv3]{
    \includegraphics[width=.3\textwidth]{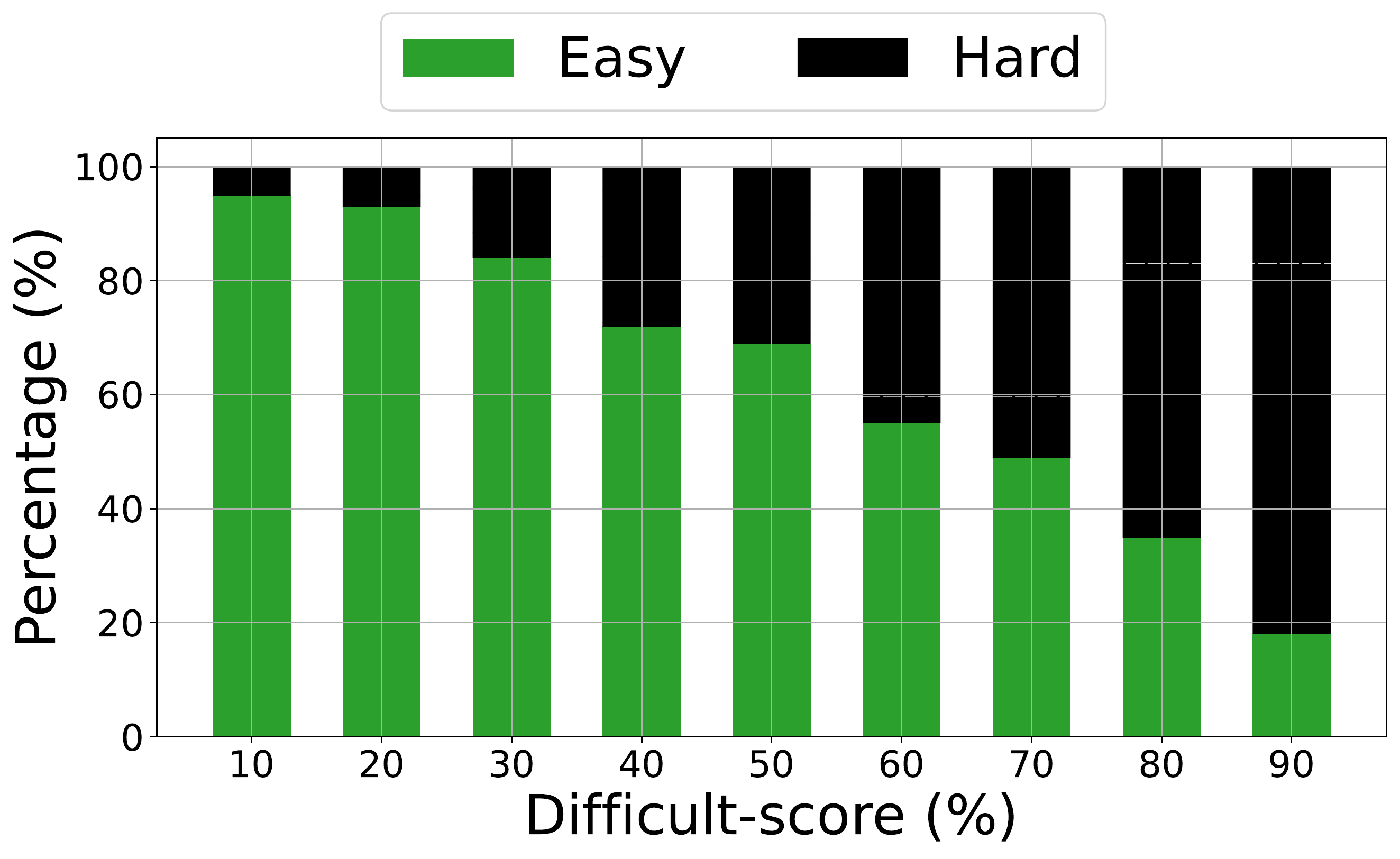}
    \label{fig:discriminator_yolov3}
    }
    \quad
    \subfigure[Faster RCNN]{
    \includegraphics[width=.3\textwidth]{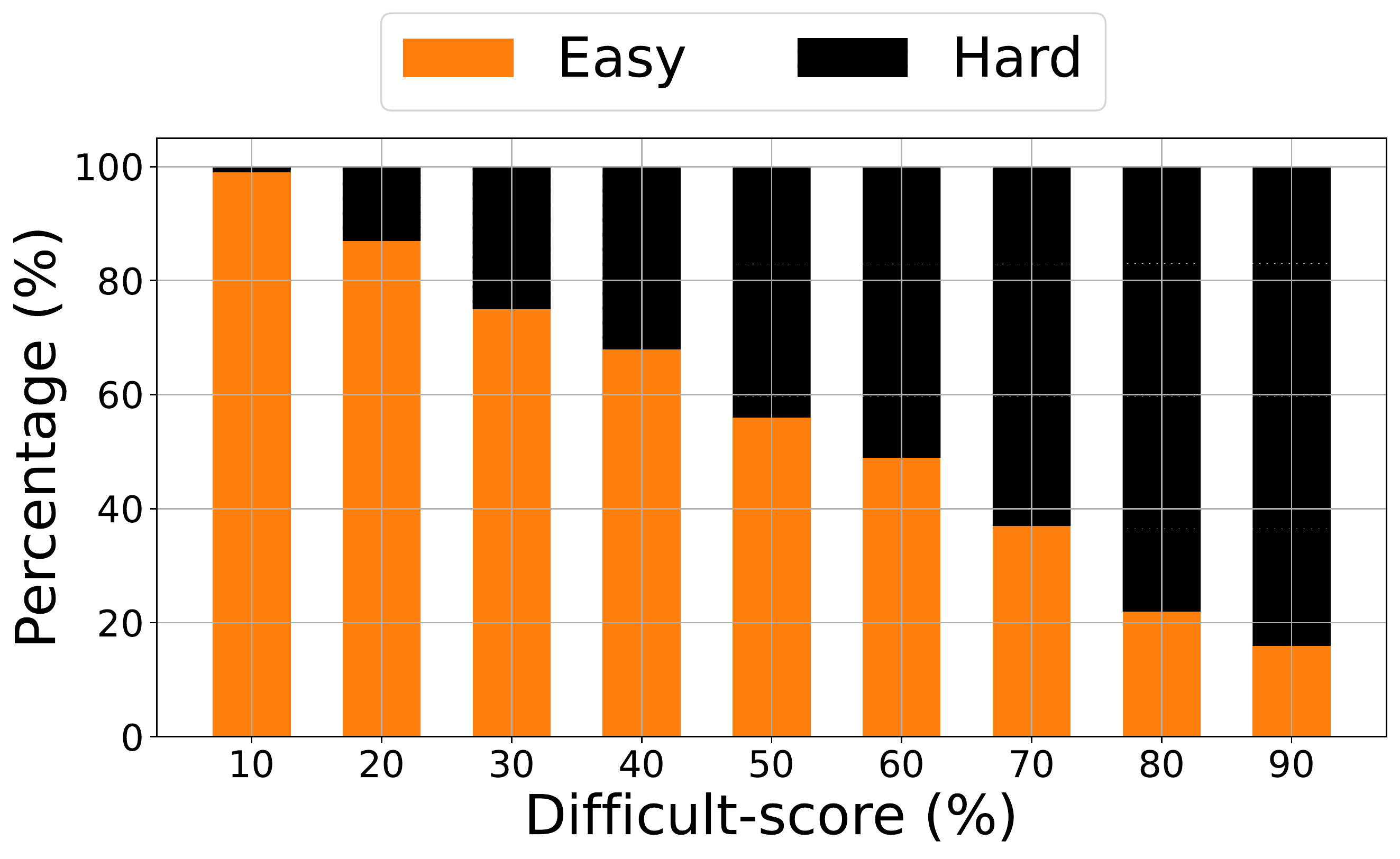}
    \label{fig:discriminator_rcnn}
    }
    \caption{An analysis on the pre-trained image-level discriminator on unseen frames.}
    \label{fig:discriminator}
\end{figure*}

\textbf{Effective pre-trained frame-level discriminator}
With an effective multi-exit GAN, \name needs a accurate and robust frame-level discriminator to predict difficult-scores for a new input frame. To examine how the pre-trained frame-level discriminator perform on the unseen data, we firstly group frames by its predicted difficult-scores and use ground-truth to get real distributions of hard samples in each group. 

As shown in Fig.~\ref{fig:discriminator}, we observe that almost all easy frames (the $1^{st}$ group) are classified correctly and more than $80\%$ hard frames in the last group are assigned to correct labels by our pre-trained frame-level discriminator. Although $15\%-18\%$ easy frames of the last group are assigned to hard samples, the number of frames in UA-DETRAC is small (about 150) and they are randomly distributed in $56,340$ testing frames. 


\begin{figure}[!t]
    \centering
    \includegraphics[width=0.85\linewidth]{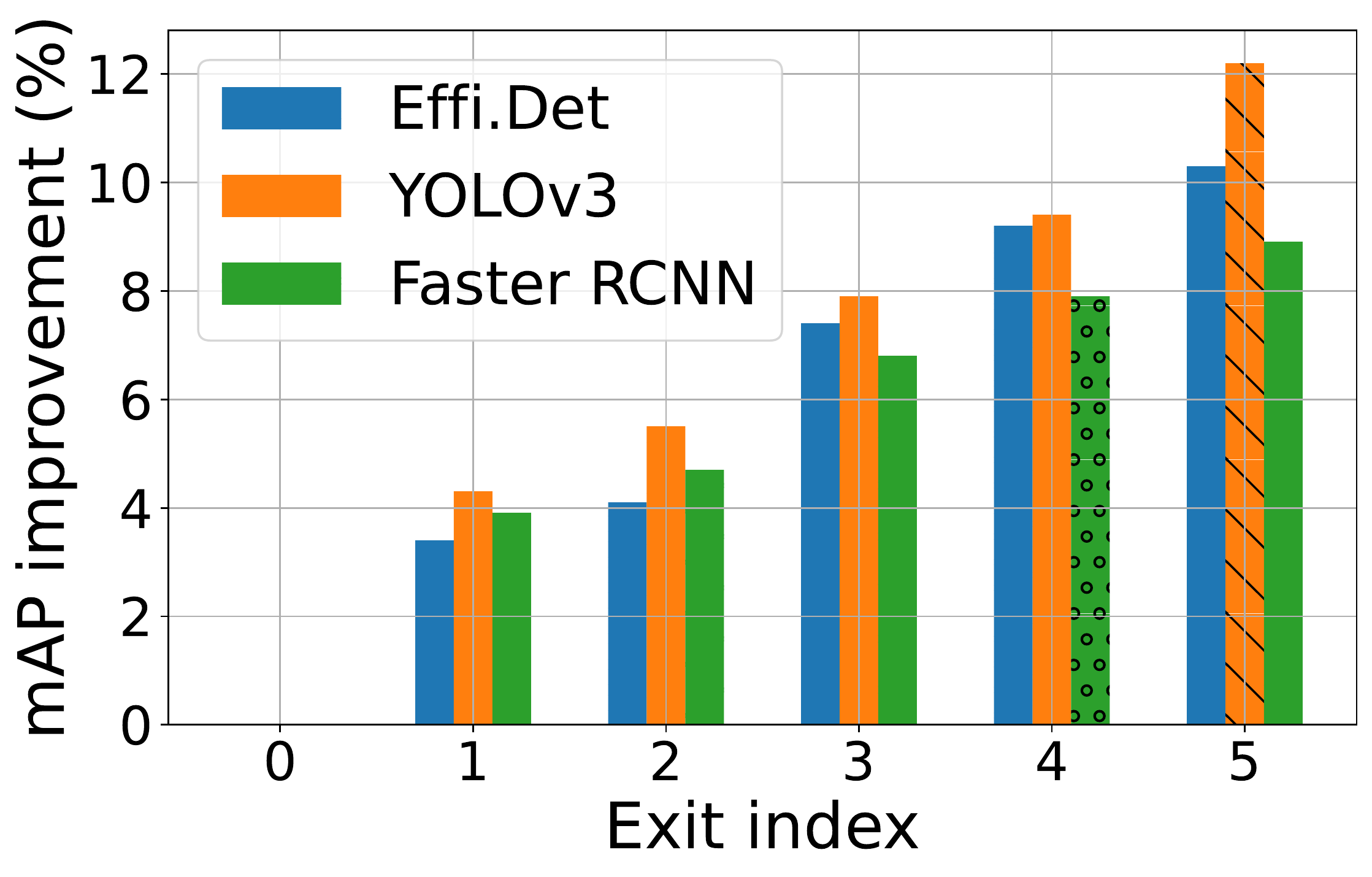}
    \label{fig:mAP_eachExit}
    \caption{Achieved mAP improvement of each exit point.}
\end{figure}

\textbf{Effectiveness of the multi-exit GAN.}
Using deeper layers of a good multi-exit $G$ on hard images should achieve higher accuracy, and our heuristic search is based on this monotonic characteristic. To verify it, we run the pre-trained multi-exit GAN on all hard frames and record overall mAP improvements for each exiting layer. We notice that the mAP can be improved monotonically via using a deeper enhancement layer, as shown in Fig.~\ref{fig:mAP_eachExit}. 

Since a deeper enhancement layer adds more semantic details of input frames, it makes the detector improve performance on the false negatives (missing objects). But on the false positives (incorrect predicted bounding boxes), image enhancement based techniques might be not working, because optimized detectors are very confident on their predicted outputs and its outputs follow a U-shaped distribution. Many bounding boxes are assigned either low ($<0.2$) or high ($>0.8$) confidences. To decrease the false positives, predictions with the high confidence should be refined, which requires the retraining on the labeled data. Thus, continuous training might be a new opportunity for \name to improve the accuracy in further.  

\begin{figure*}[!th]
    \centering
    \subfigure[EfficientDet-D0]{
        \begin{minipage}{0.32\textwidth}
            \centering
            \includegraphics[width=\textwidth]{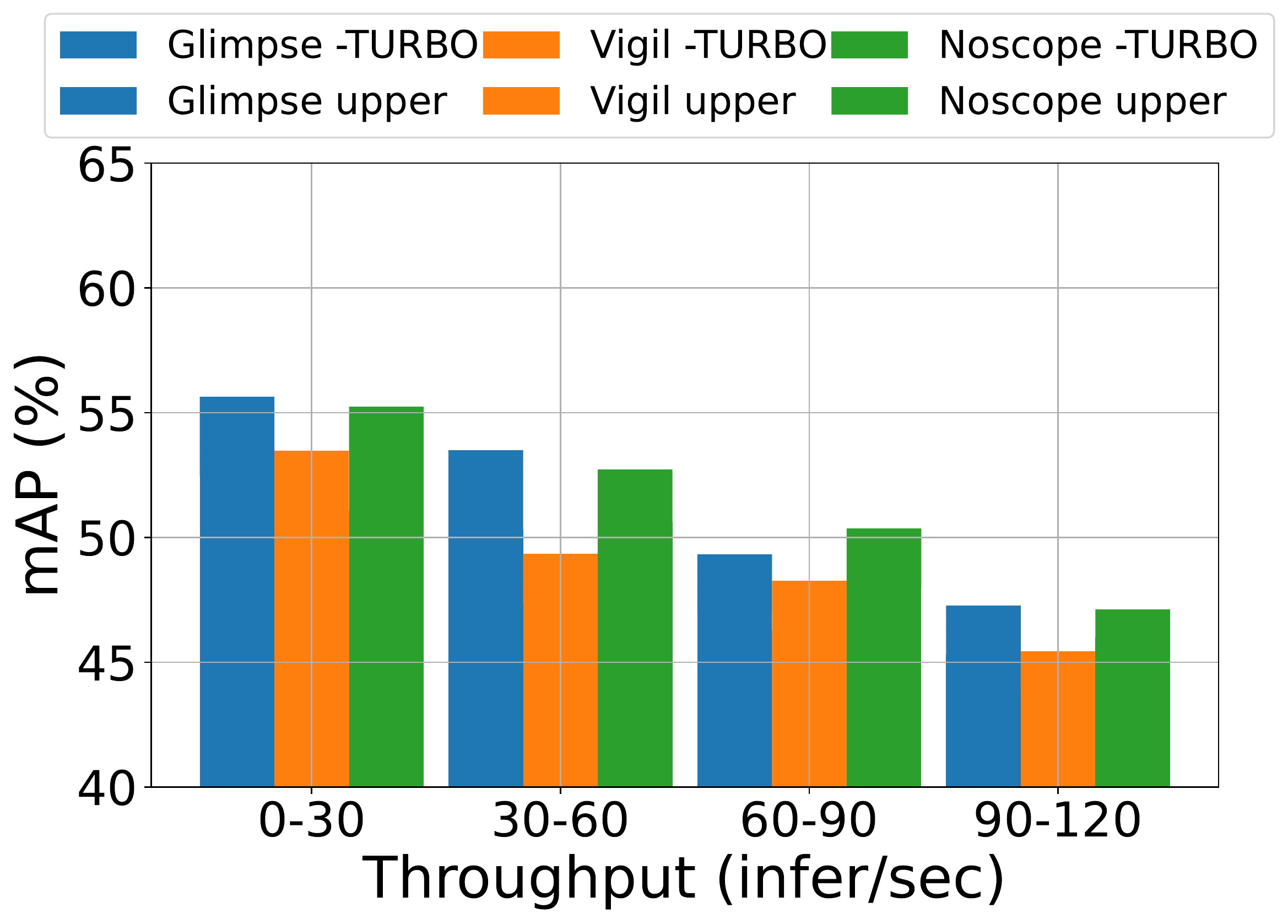}
            \label{fig:noscope_search_map}
        \end{minipage}
    }
    \subfigure[YOLOv3]{
    \begin{minipage}{0.32\textwidth}
        \centering
        \includegraphics[width=\textwidth]{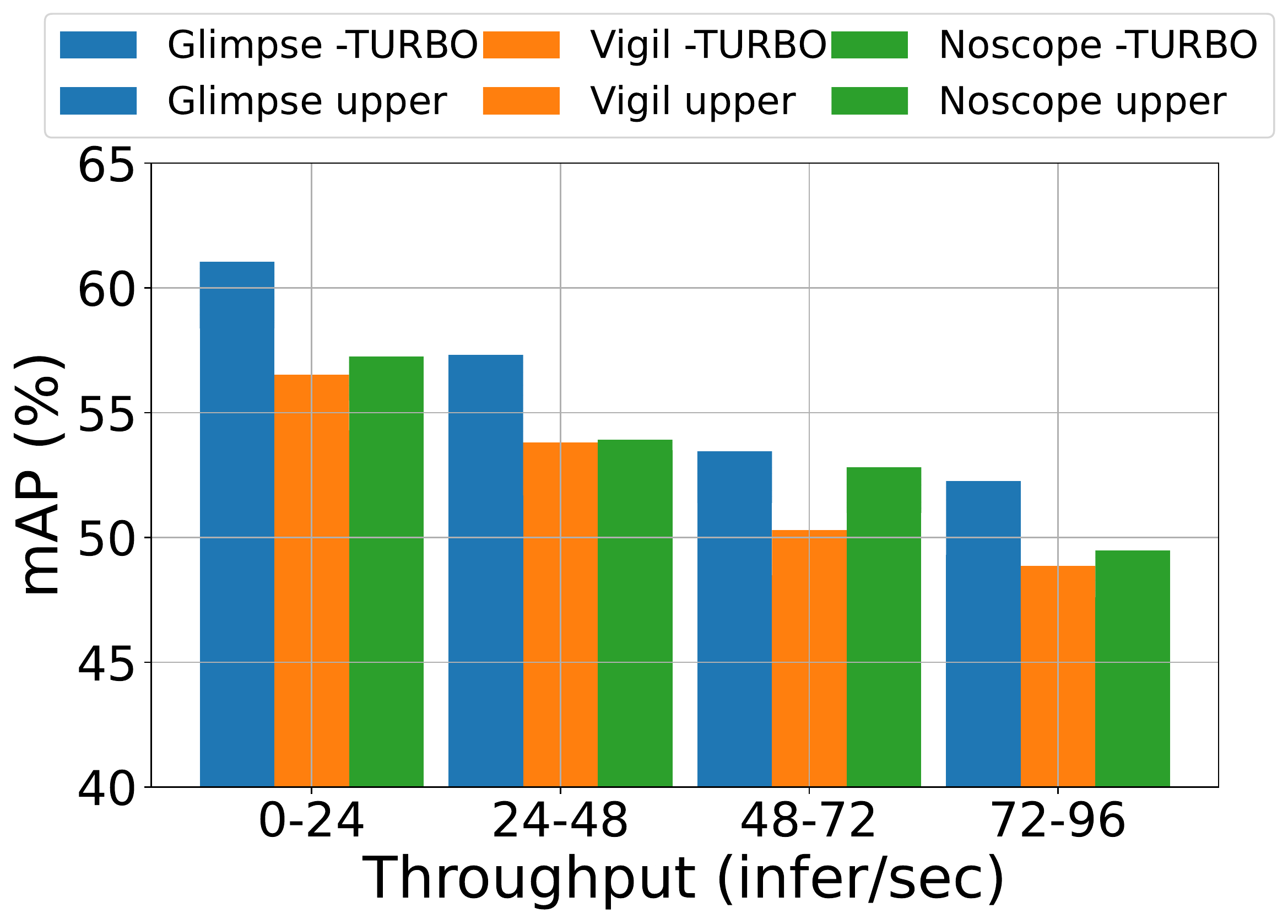}
        \label{fig:vigil_search_map}
    \end{minipage}
    }
    \subfigure[Faster RCNN]{
    \begin{minipage}{0.32\textwidth}
        \centering
        \includegraphics[width=\textwidth]{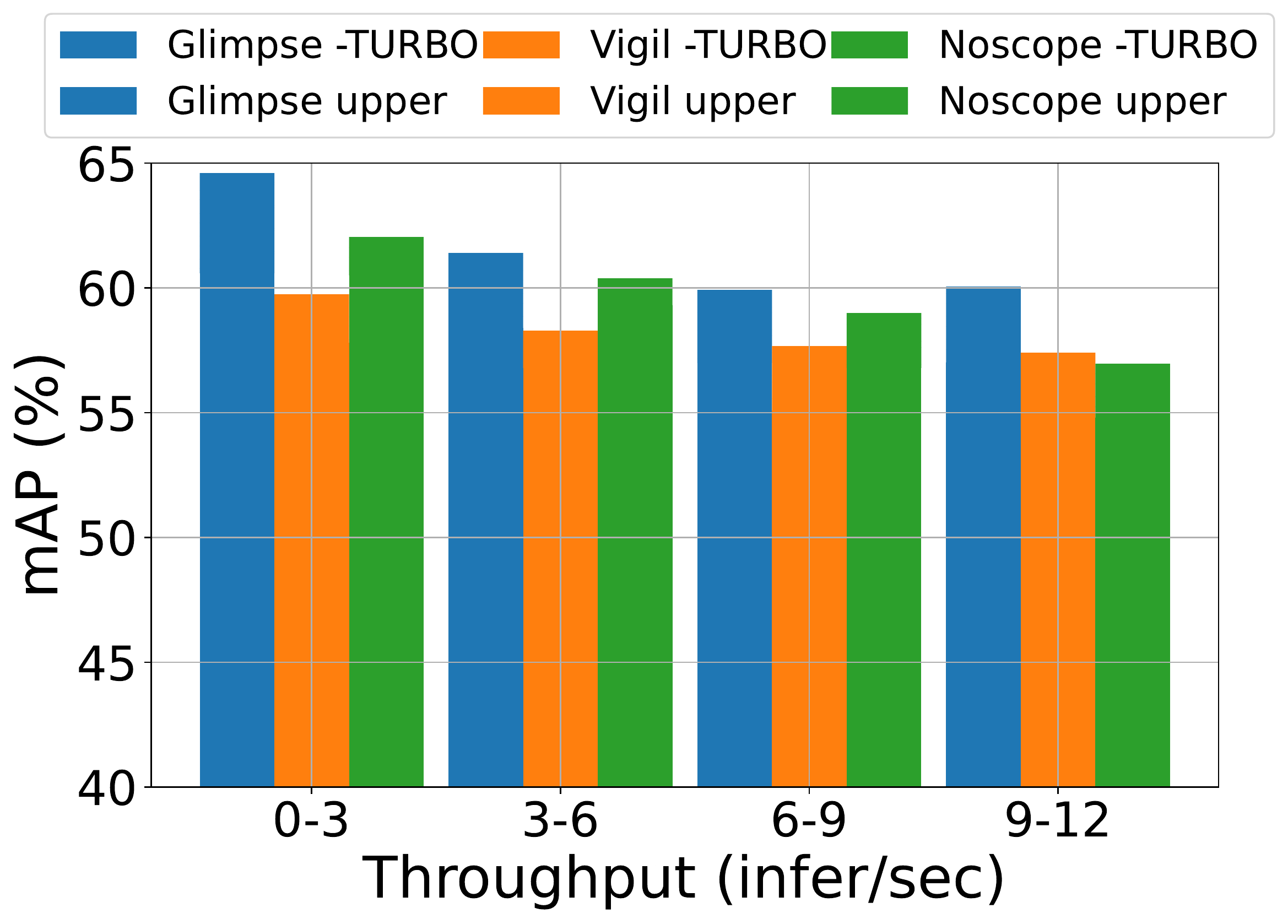}
        \label{fig:glimpse_search_map}
    \end{minipage}
    }
    \quad 
    \subfigure[EfficientDet-D0]{
    \begin{minipage}{0.32\textwidth}
        \centering
        \includegraphics[width=\textwidth]{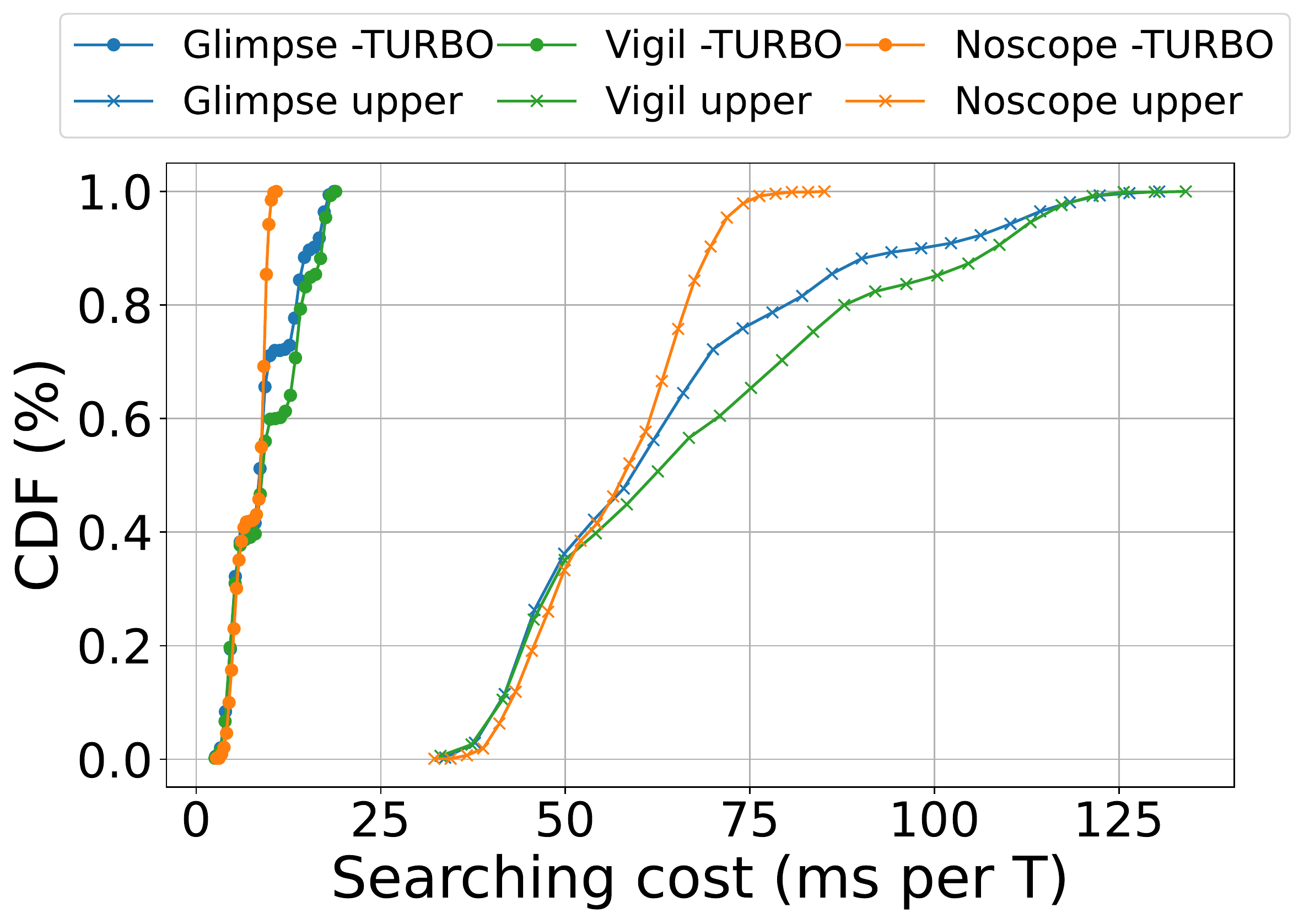}
        \label{fig:glimpse_search_cost}
    \end{minipage}
    }
    \subfigure[YOLOv3]{
        \begin{minipage}{0.32\textwidth}
            \centering
            \includegraphics[width=\textwidth]{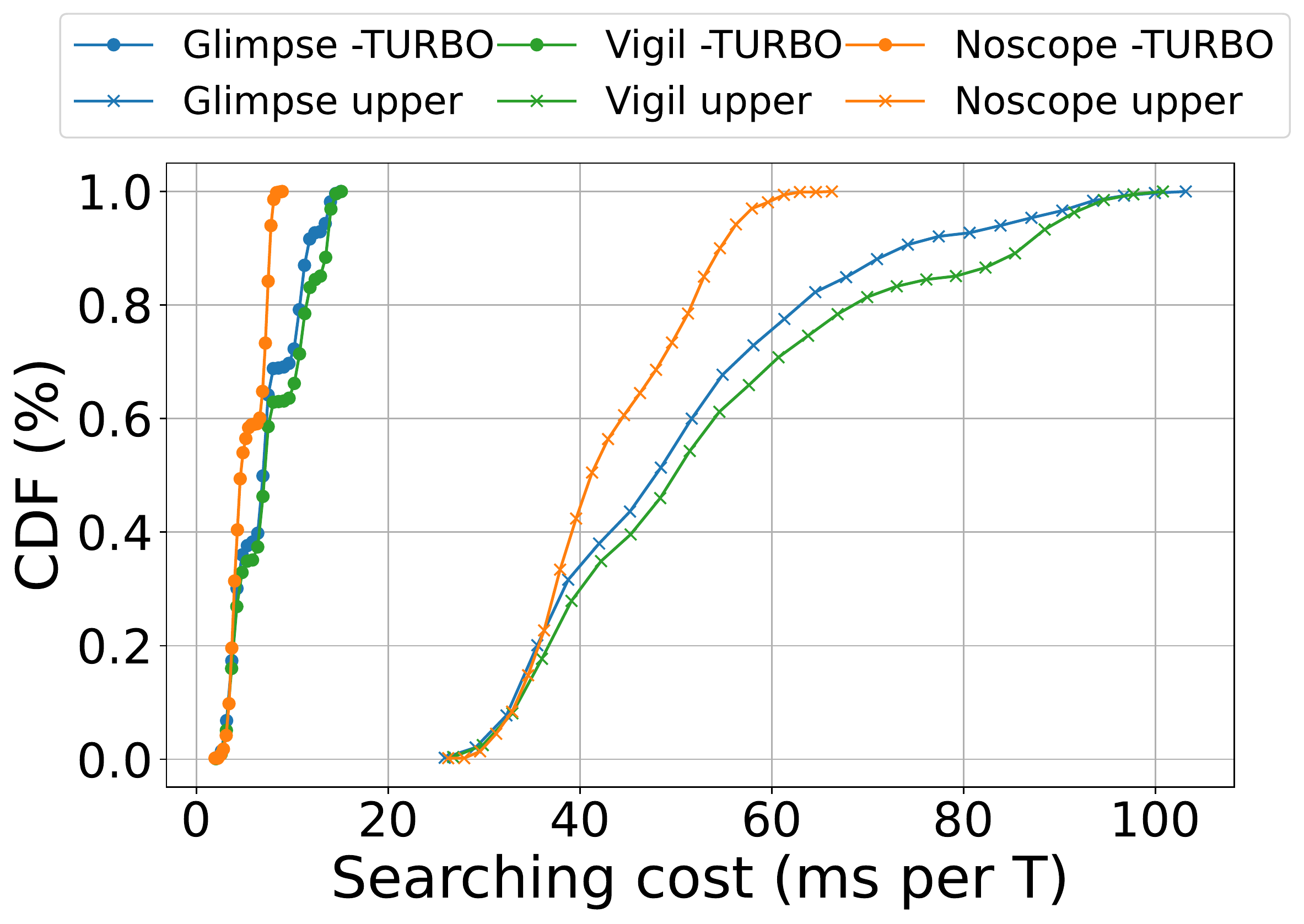}
            \label{fig:vigil_search_cost}
        \end{minipage}
    }
    \subfigure[EfficientDet-D0]{
    \begin{minipage}{.32\textwidth}
        \centering
        \includegraphics[width=\textwidth]{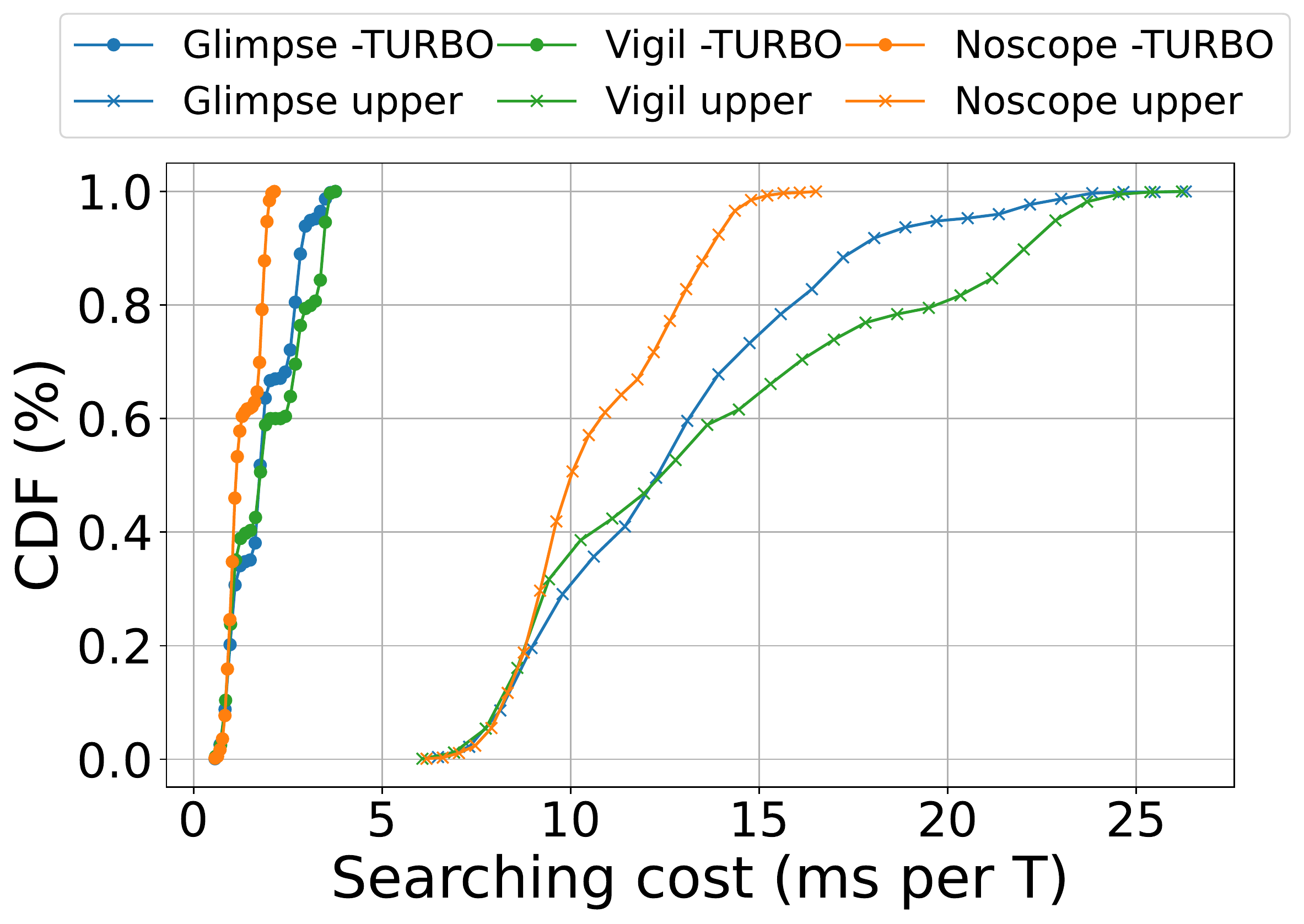}
        \label{fig:noscope_search_cost}
    \end{minipage}
    }
    \caption{Accuracy gains and search costs of the \name scheduler, compared with the brute-force searching based scheduler (denoted as \emph{upper}).}
    \label{fig:optimal_scheduler}
\end{figure*}

\subsection{Evaluation of Adaptive Enhancement Scheduling}
In the end, we evaluate our scheduler. Because our solution is based on the heuristic idea and achieves sub-optimal trade-off between the accuracy and the latency, we compare \name with a brute-force search method, notated as \emph{upper}. As shown in Fig.~\ref{fig:optimal_scheduler}, \name is lower than the upper-bound method by average $3.26\%$, $1.95\%$ and $3.63\%$ mAP for EfficientDet-D0, YOLOv3 and Faster RCNN respectively. To measure searching cost on different scheduling methods, we present the distribution of execution time for three detectors on Glimpse, Vigil and Noscope in Fig.~\ref{fig:glimpse_search_cost}, Fig.~\ref{fig:vigil_search_cost} and Fig.~\ref{fig:noscope_search_cost} respectively. We find that even though a brute-force search can achieve the best accuracy but its latency of searching is much larger than \name. Thus, our scheduling algorithm is an efficient and effective method because it can find a enough good solution within about $20ms$.


\section{Related work}
\textbf{Edge Video Analytics Pipelines.}
Edge video analytics systems have been widely deployed and have became the solution to many large-scale safety and management tasks. Driven by advances in machine learning and hardware acceleration, most systems~\cite{zhang2017live,noscope2017,awstream2018,glimpse2015,vigil2015,jiang2018chameleon,hung2018videoedge,Reducto,dds2018,eaar2019,weg2017} are follow an edge-cloud architecture~\cite{measurement2021xiao} where cameras are responsible for processing simple tasks (\eg video compression and temporal filtering) and edge servers maintain a deep neural network to provide accurate analysis on input videos. However, inference accuracy is often limited by a unstable network and resource-constraint edge servers. To balance the inference accuracy and resource (compute/network) cost, many cascade video analytics pipelines are proposed. 

They leverage video processing heuristics to design three pruning methods on cameras and save large unnecessary compute and transmission costs. The first~\cite{zhang2017live,noscope2017,awstream2018,glimpse2015,vigil2015,jiang2018chameleon,hung2018videoedge,Reducto,eaar2019} is based on temporal consistency of videos and remove similar frames by inter-frame difference~\cite{noscope2017,glimpse2015}. Although they are effective to reduce transmission cost, a cheap tracking model is required to be deployed on cameras for inference on filtered frames. To further reduce communication cost, spatial~\cite{zhang2017live,awstream2018,vigil2015,jiang2018chameleon} and model~\cite{zhang2017live,noscope2017,jiang2018chameleon,weg2017} pruning methods are also widely used in VAPs. Both methods need to deploy a cheap deep neural network on edge devices and use it to select region-of-interests (RoIs) or uncertain frames for server's inference. In practice, three pruning methods are often used together (\eg Noscope~\cite{noscope2017} integrates temporal and model pruning to filter video frames). Based on this cascade design, edge servers only need to process few frames in a video. However, a over-provisional compute resource is often assigned because it can meet all requests' requirements. Thus, how to leverage this existing dynamic idle compute resources to improve inference is ignored by current video analytics systems. To bridge this gap, we design a detector-specific multi-exit GAN to enhance hard samples adaptively for higher overall accuracy.

\textbf{Image Enhancement.}
Image enhancement is a well-studied problem in low-level computer vision (CV) tasks and is also named image restoration~\cite{Kaur2014ImageR}. In many benchmarks~\cite{Ignatov_2018_ECCV_Workshops,young-etal-2014-image}, they are explicit grouped by the corresponding image noise: image deblurring~\cite{Zamir2021MPRNet}, image deraining~\cite{Zamir2021MPRNet}, image denosing~\cite{liang2021swinir}, image dehazing~\cite{qin2020ffa}, relight~\cite{jiang2021enlightengan,Zero-DCE2020} and super-resolution~\cite{liang2021swinir}. It aims to restore raw images from images mixed by noises and is always seen as a data prepossessing step for downstream CV tasks. Because almost vision models (\eg YOLOv3~\cite{yolov3} and EfficientDet-v0~\cite{efficientdet2020}) are pre-trained on cleaned images (\eg Microsoft COCO~\cite{coco2014}), their accuracy are easy to be degraded by natural image noise. For example, vehicle detectors are often degraded by image blurring when cars speed up. Besides, existing benchmarks only provide labeled training data for a single noise instead of mixed noise. Thus, we need to download different pre-trained models for processing different image noise. But in real world applications, an video may contain more than two or three noises. For instance, traffic videos collected in a rush hour may contain blurring and low-light cases. Thus, using them in VAPs is not easy because it requires VAPs select suitable models on any frames. Fortunately, we find that adversarial training can provide not only image generator but also image discriminator, which is highly matching VAP's need on frames' selection. Thus, we choose to integrate existing VAPs with a GAN-based image enhancement model~\cite{jiang2021enlightengan}.

\section{Discussion and Future works}
We propose \name to improve the inference accuracy via executing an GAN-based enhancement module by harvesting the dynamic idle compute resource for edge VAPs. We notice that the achieved accuracy gain highly relates to the filtering methods. For example, the model-based filtering always brings the higher accuracy improvements than the temporal filtering. Thus, an interesting solution is ot use an adaptive hyper-parameter to adjust the filtering rate and the accuracy gain together. To scale \name to spatial filtering based VAPs, we need to develop a region-level GAN to enhance semantic details of hard RoIs. In the future, we would put more engineering efforts to scale \name to VAPs with more advanced filtering modules.

Although our multi-exit GAN is an effective tool to enhance hard samples within a latency constraint, over-fitting issues still remain in model-aware adversarial training. GAN's generalization on new hard images are limited. Motivated by continuous training algorithms, we would like to preserve the GAN's performance through fine-tuning on the new testing data.

\section{Conclusion}
In this paper, we propose \name, an opportunistic image enhancement framework which takes advantages of the over-provisioned GPU resources at runtime to improve the overall video analytics accuracy. \name first designs a task-specific GAN and trains it with the model-aware adversarial training strategy. Such a method allows the GAN to intelligently identifies model-specific hard samples and applies enhancements at various granularity. At runtime, an enhancement execution scheduler is developed to assign the most suitable enhancement level to each image to achieve the best overall accuracy within a given resource availability. We evaluate \name on a real-world traffic video dataset with three canonical video analytics pipelines. \name improves the absolute mAP by $9.02\%$, $11.34\%$ and $7.27\%$ on average for EfficientDet-D0, YOLOv3 and Faster RCNN, respectively.

\bibliographystyle{ACM-Reference-Format}
\bibliography{reference}

\clearpage

\appendix
\section*{APPENDIX}

\section{A measurement study with NoScope}
\label{sec:measurement_noscope}
\begin{figure}[!t]
        \centering
        \includegraphics[width=0.95\linewidth]{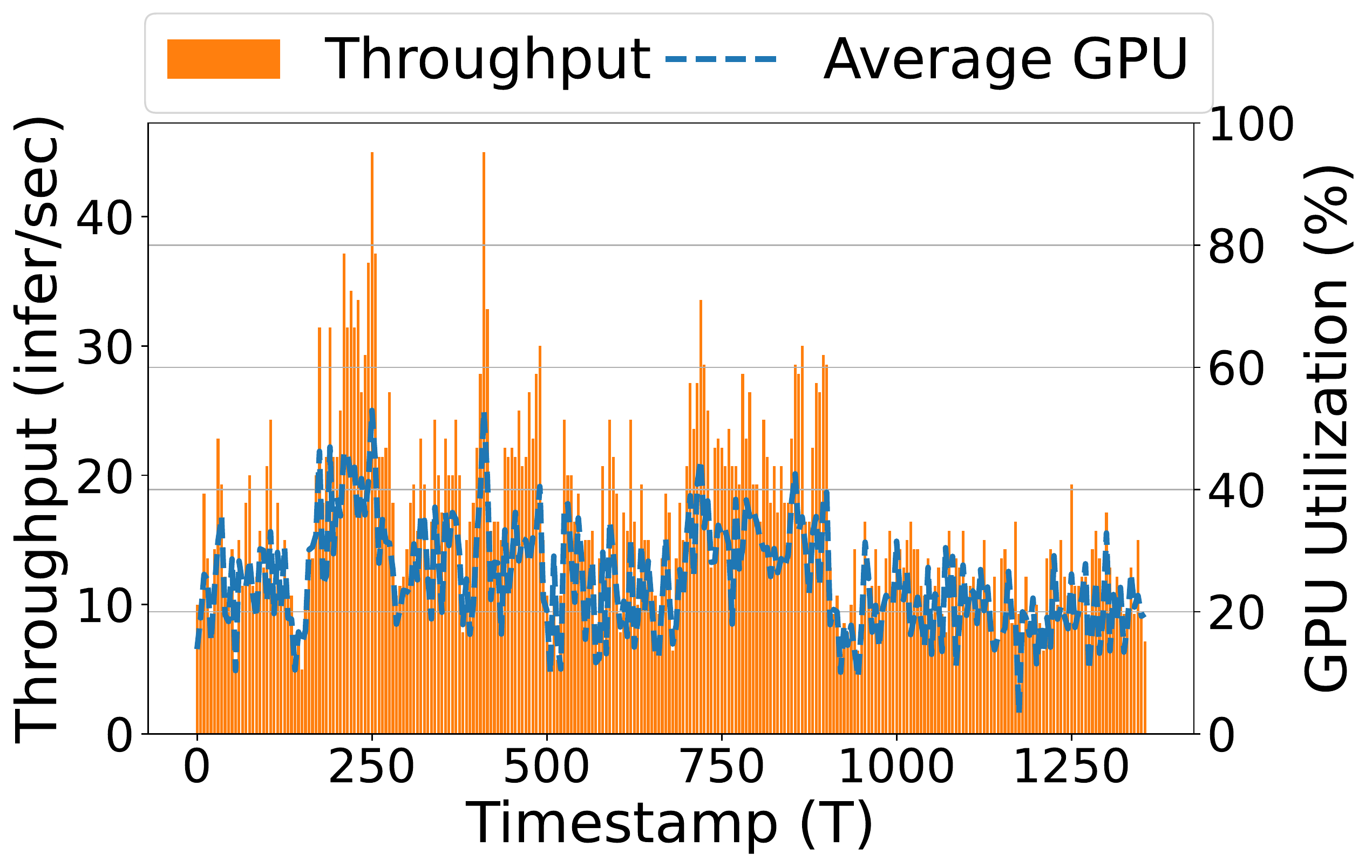}
        \caption{Dynamic video analytics workloads with on an edge device shared by multiple streams.}
        \label{fig:measurement_noscope}
\end{figure}

Using a same setting (EfficientDet-D0 and 4 video streams), we execute Noscope~\cite{noscope2017} on UA-DETRAC~\cite{CVIU_UA-DETRAC} and show the actual workload in Fig.~\ref{fig:measurement_noscope}. Like Vigil~\cite{vigil2015} and Glimpse~\cite{glimpse2015}, GPU throughput of Noscope varies greatly over time, from 6 infer/sec to 44 infer/sec. It is interesting to note that GPU throughput of Noscope is much lower than other two pipelines and it stays below 20 infer/sec for 83.1\% time. It is because that a cascade design of temporal and model pruning methods in Noscope makes less frames for the downstream heavyweight object detection model.

\section{Enhancement profiling}
\label{sec:enhancement_profiling}
\begin{figure}[!t]
        \centering
        \includegraphics[width=0.95\linewidth]{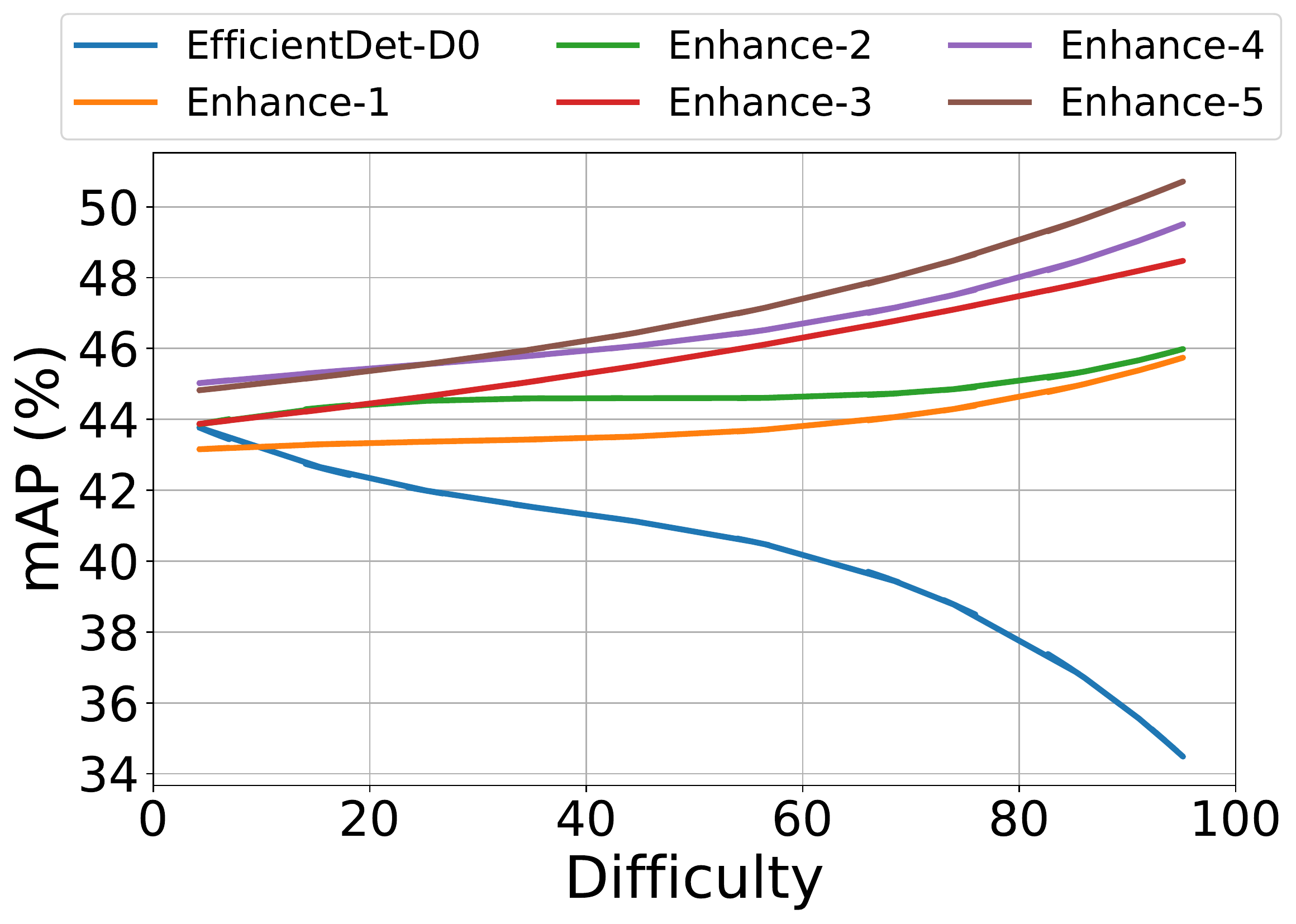}
        \caption{mAP of executing the $k_{th}$ level enhancement on UA-DETRAC with the different difficulty ($\theta$).}
        \label{fig:enhancement_profiling}
\end{figure}

Enhancement profiling aims to profile the accuracy gain of the multi-exit GAN and the expected inference latency of executing the $k_{th}$ level enhancement. Fig.~\ref{fig:enhancement_profiling} shows the mAP of running the $k_{th}$ level enhancement on UA-DETRAC~\cite{CVIU_UA-DETRAC} with the different difficulty of $\theta$. Obviously, using a higher enhancement makes a larger mAP improvement. In specific, average mAP improvement of the $5_{th}$ level enhancement is $6.15\%$ higher than the $1_{st}$ level enhancement. In addition, a frame with a larger difficulty ($\theta$) would be enhanced better by a higher enhancement. For example, average mAP improvement of running the $5_{th}$ level enhancement on frames with the difficulty ($90-100$) is $5.54\%$ higher than frames with the difficulty ($80-90$).

\section{End-to-End Evaluation}
\label{sec:e2e_res_appendix}
\begin{figure*}[!ht]
\subfigure[EfficientDet-D0]{
\begin{minipage}{0.32\textwidth}
        \centering
        \includegraphics[width=\textwidth]{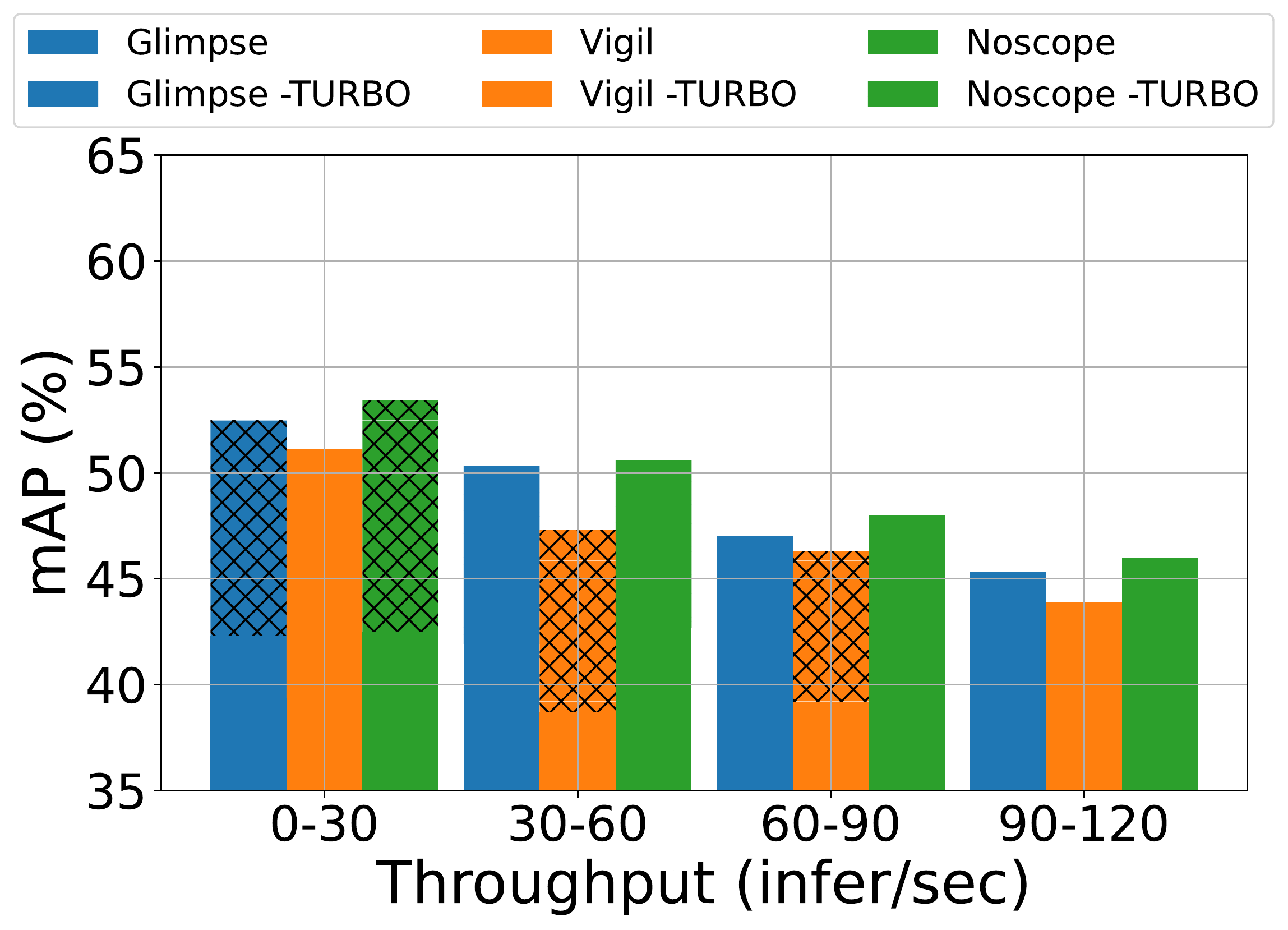}
        \label{fig:map_throughput_detrac_d0}
\end{minipage}}
\hfill
\subfigure[YOLOv3]{
\begin{minipage}{0.32\textwidth}
        \centering
        \includegraphics[width=\textwidth]{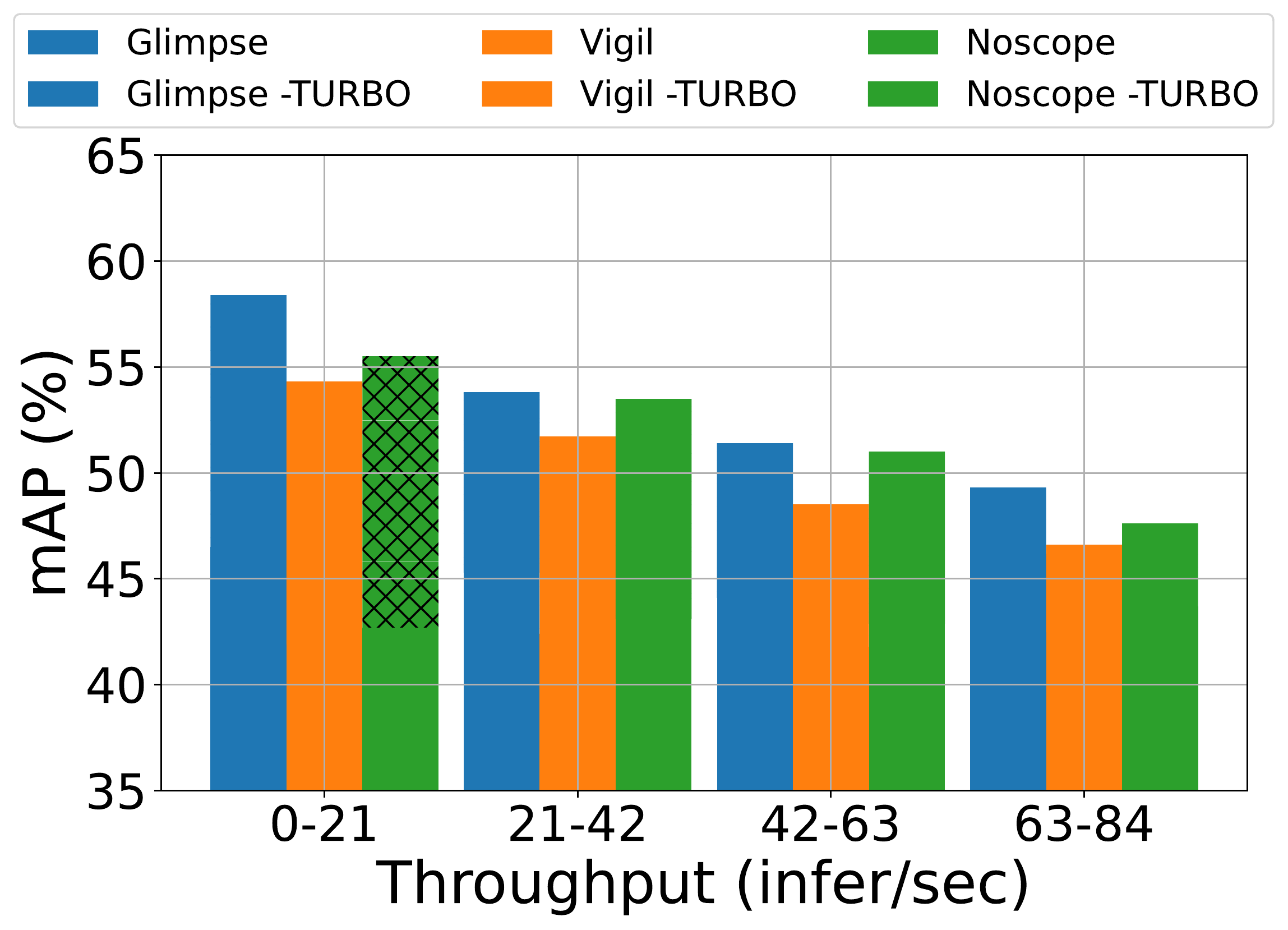}
        \label{fig:map_throughput_detrac_yolov3}
\end{minipage}}
\hfill
\subfigure[Faster RCNN]{
\begin{minipage}{0.32\textwidth}
    \centering
    \includegraphics[width=\textwidth]{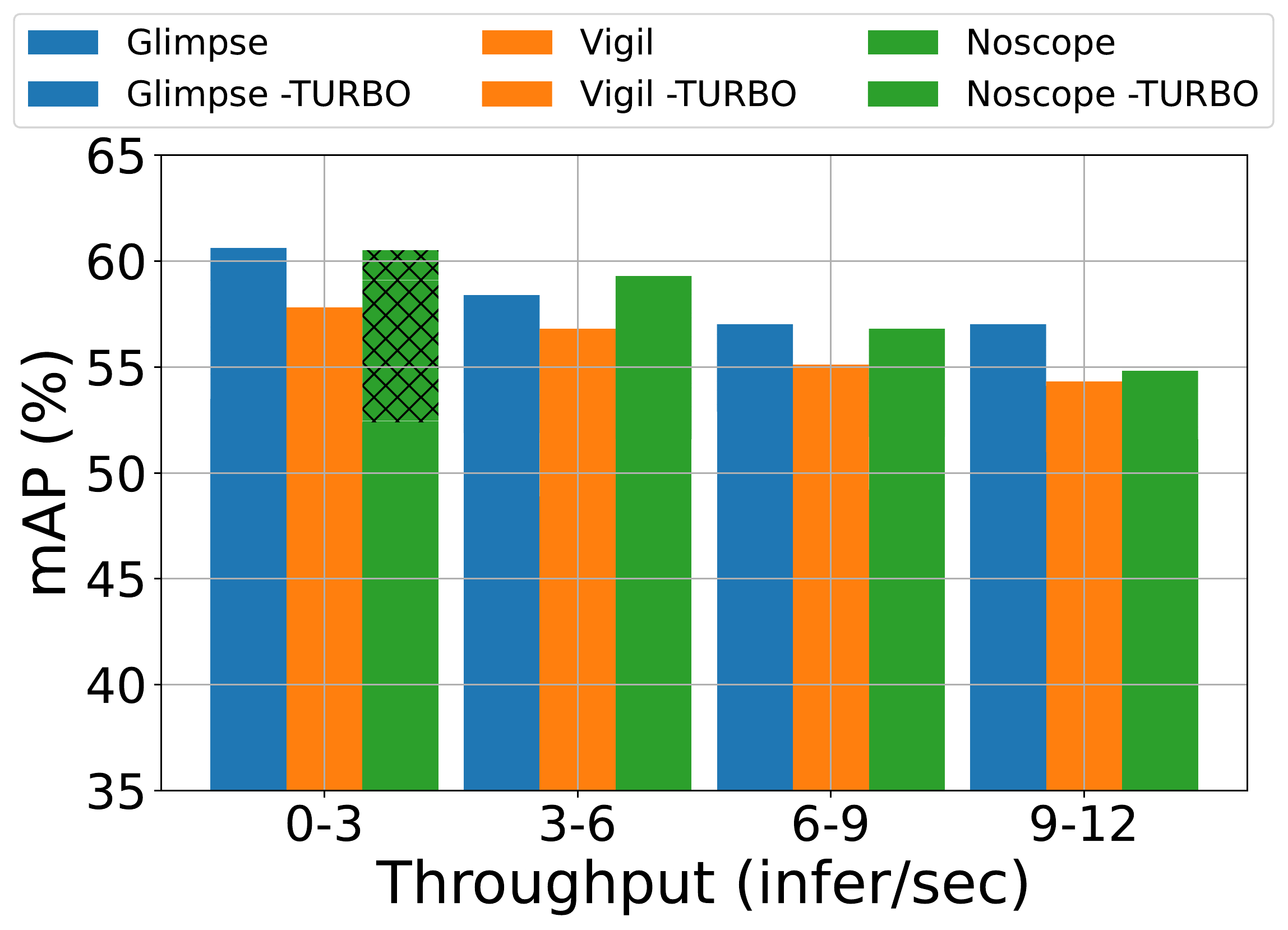}
    \label{fig:map_throughput_detrac_rcnn}
\end{minipage}}
\caption{mAP of the selected VAPs and the enhanced VAPs by Turbo with three object detectors under different throughput, using T4 GPU on UA-DETRAC.}
\label{fig:map_throughput_detrac}
\end{figure*}

\begin{figure*}[!ht]
\subfigure[EfficientDet-D0]{
\begin{minipage}{0.32\textwidth}
        \centering
        \includegraphics[width=\textwidth]{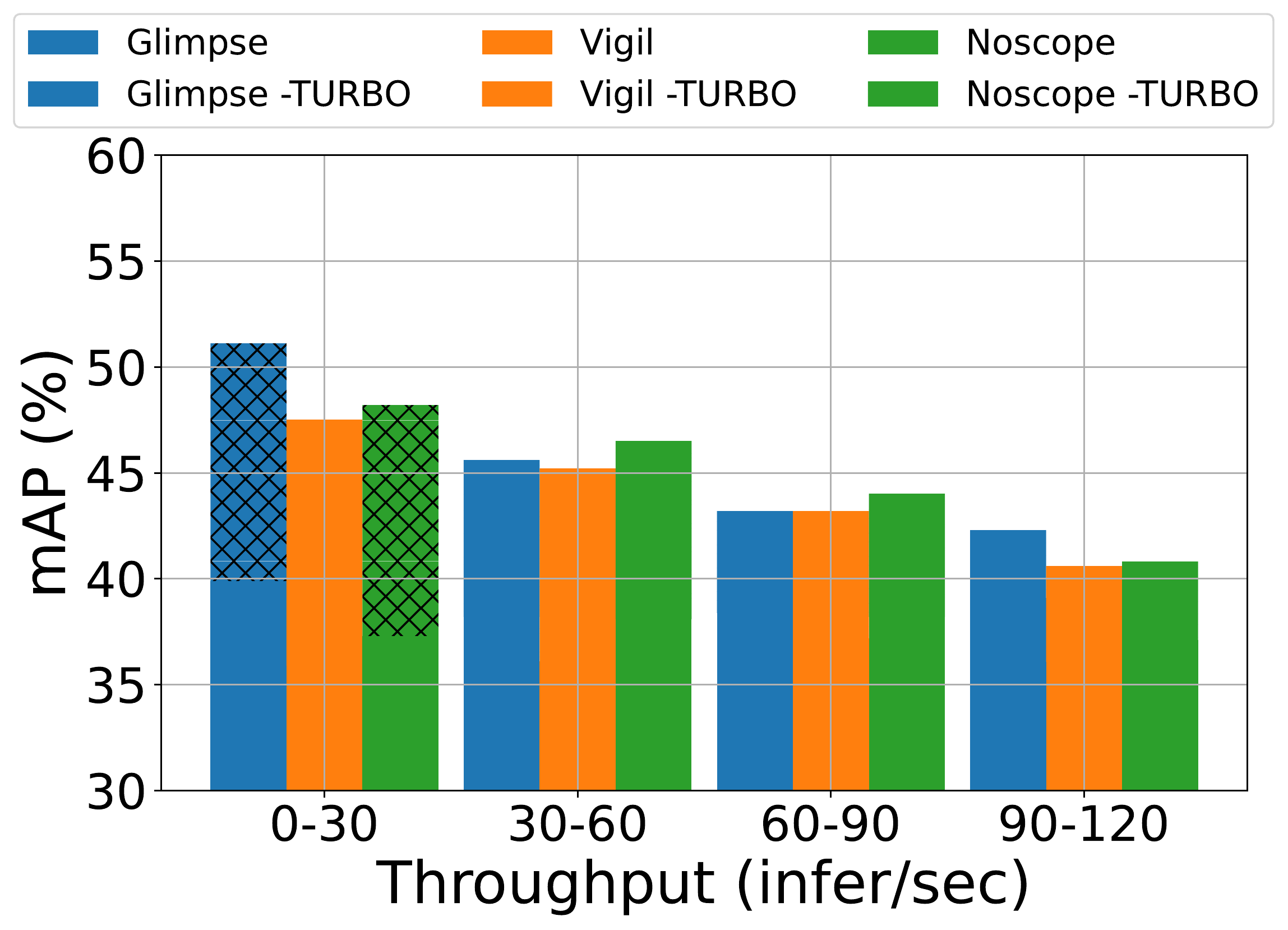}
        \label{fig:map_throughput_aicity_d0}
\end{minipage}}
\hfill
\subfigure[YOLOv3]{
\begin{minipage}{0.32\textwidth}
        \centering
        \includegraphics[width=\textwidth]{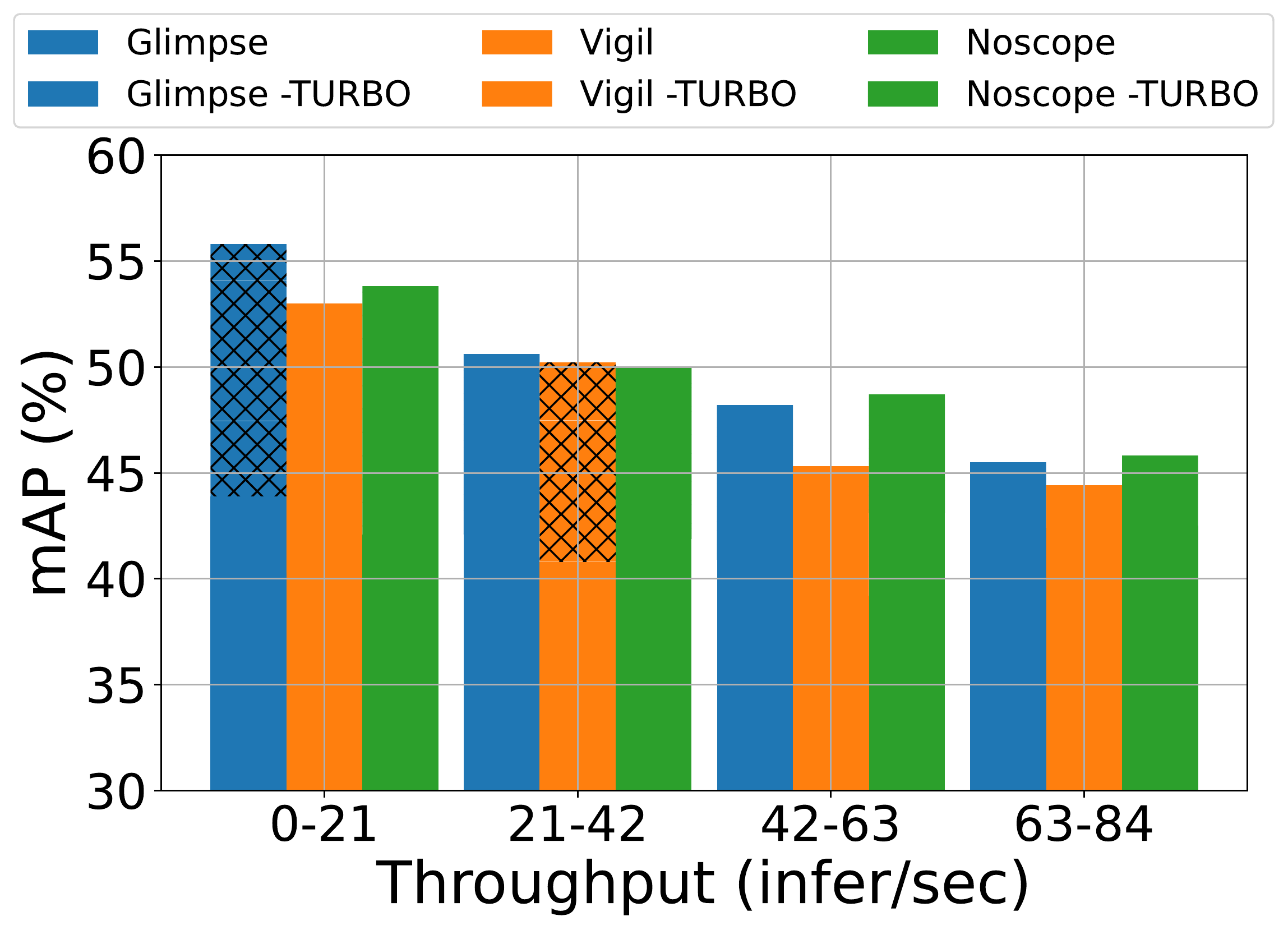}
        \label{fig:map_throughput_aicity_yolov3}
\end{minipage}}
\hfill
\subfigure[Faster RCNN]{
\begin{minipage}{0.32\textwidth}
    \centering
    \includegraphics[width=\textwidth]{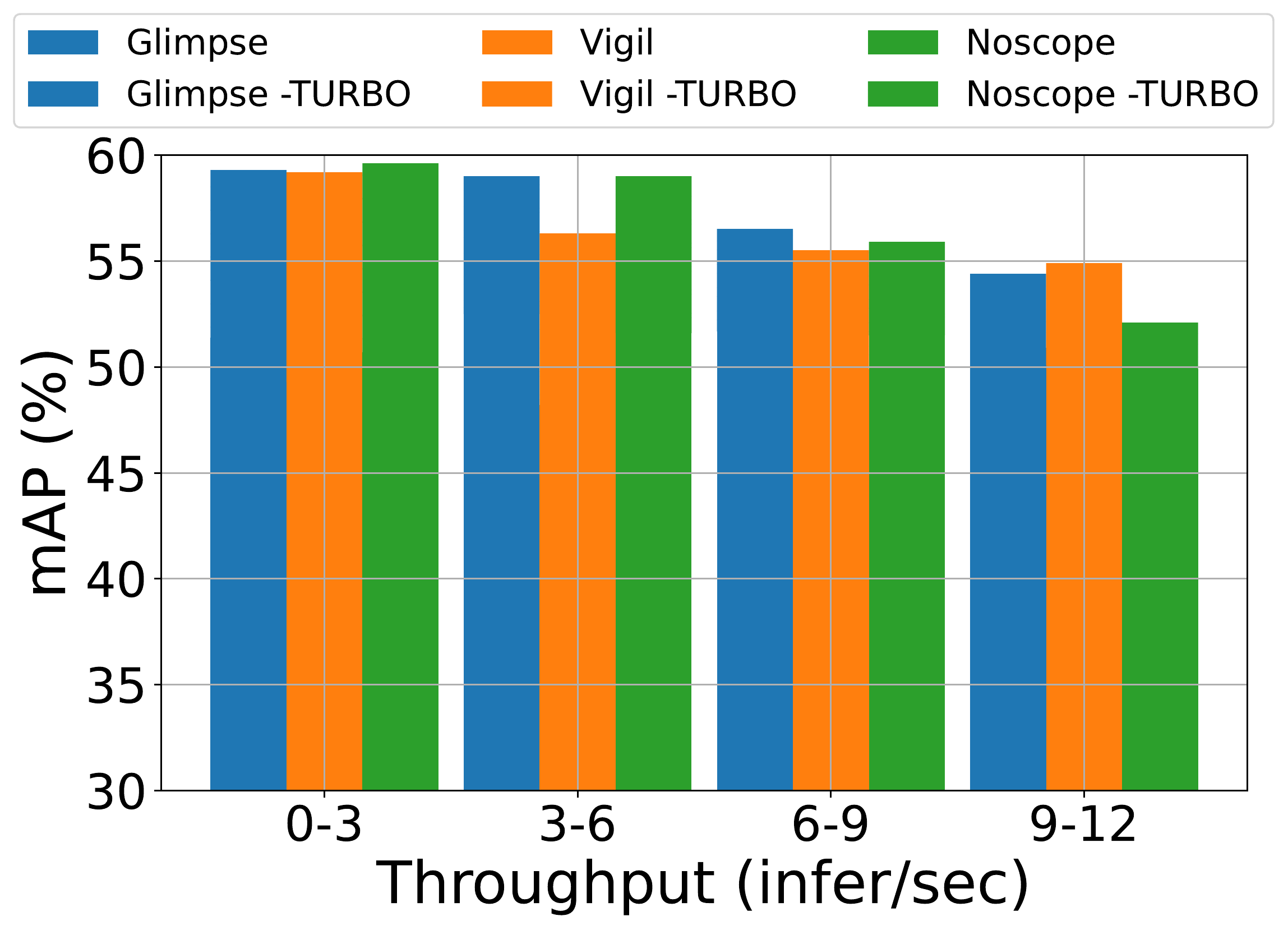}
    \label{fig:map_throughput_aicity_rcnn}
\end{minipage}}
\caption{mAP of the selected VAPs and the enhanced VAPs by Turbo with three object detectors under different throughput, using T4 GPU on AICity.}
\label{fig:map_throughput_aicity}
\end{figure*}

To evaluate the overall performance of \name completely, we report its overall mAP on AICity~\cite{Tang19CityFlow} with three object detectors using V100. In addition, we report performance on UA-DETRAC~\cite{CVIU_UA-DETRAC} and AICity~\cite{Tang19CityFlow} over different throughput with three object detectors in Figure~\ref{fig:map_throughput_detrac} and Figure~\ref{fig:map_throughput_aicity}, respectively.

\begin{figure}[!t]
        \centering
        \includegraphics[width=0.95\linewidth]{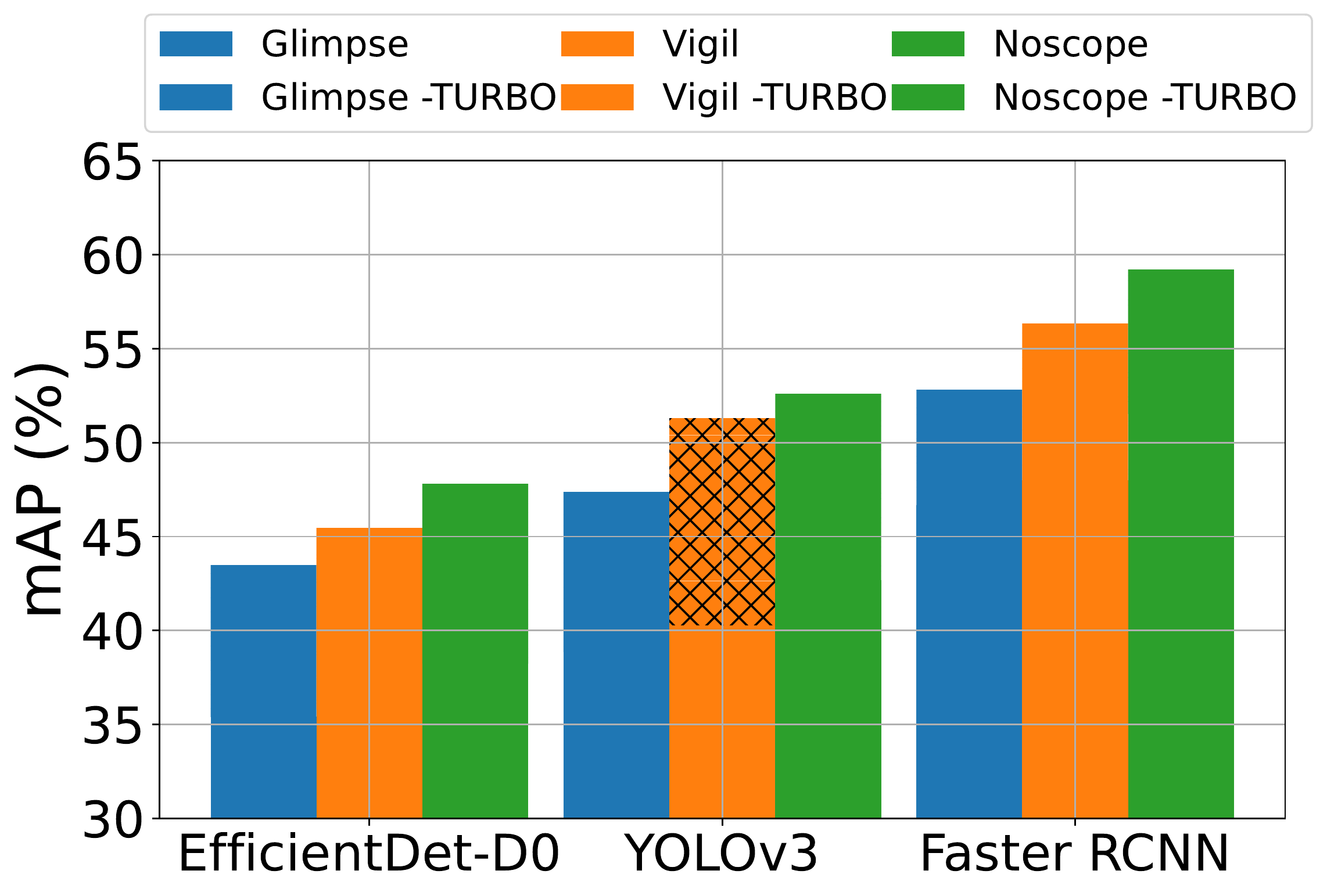}
        \caption{Overall mAP on AICity of the selected VAPs and the enhanced VAPs by \name, with different object detectors, on V100 GPU.}
        \label{fig:overall_v100_aicity}
\end{figure}

\textbf{Performance over different datasets.} As shown in Fig.~\ref{fig:overall_v100_aicity}, similar mAP improvements of three object detectors can be achieved by \name using V100. Overall the average mAP improvments of Glimpse, Vigil and NoScope on AICity are 7.03\%, 9.73\% and 8.62\%. 

\textbf{Performance over different throughput.} As described above, throughput is the real arriving rate of video streams, it represents the number of frames the VAP need to process per second. According to our evaluation, the maximum processing capacity of T4 for EfficientDet-D0, YOLOv3 and Faster RCNN are 120, 84, 12. Fig.~\ref{fig:map_throughput_detrac} and Fig.~\ref{fig:map_throughput_aicity} illustrate the
obtained mAP of the enhanced VAPs under different throughout on UA-DETRAC and AICity, respectively. \name performs better when the throughput is low. Especially on YOLOv3, 13.75\% and 11.3\% absolute mAP improvement can be obtained for three VAPs when the throughput is lower than 21 infer/sec.

\end{document}